\PassOptionsToPackage{dvipsnames, table}{xcolor}
\documentclass{article} % For LaTeX2e
\usepackage{iclr2026_arxiv,times}
\iclrfinalcopy

\usepackage[utf8]{inputenc} % allow utf-8 input
\usepackage[T1]{fontenc}    % use 8-bit T1 fonts
\usepackage{hyperref}       % hyperlinks
\usepackage{url}            % simple URL typesetting
\usepackage{booktabs}       % professional-quality tables
\usepackage{amsfonts}       % blackboard math symbols
\usepackage{nicefrac}       % compact symbols for 1/2, etc.
\usepackage{microtype}      % microtypography
\usepackage{xcolor}         % colors
\usepackage[table, dvipsnames]{xcolor}

\usepackage{amsmath,amsfonts,bm}

\def\eqref#1{equation~\ref{#1}}
\def\1{\bm{1}}

\DeclareMathAlphabet{\mathsfit}{\encodingdefault}{\sfdefault}{m}{sl}
\SetMathAlphabet{\mathsfit}{bold}{\encodingdefault}{\sfdefault}{bx}{n}

\DeclareMathOperator*{\argmax}{arg\,max}

\usepackage{times}
\usepackage{latexsym}
\usepackage{amsmath}
\usepackage{amsfonts}
\usepackage{graphicx}
\usepackage{amssymb}
\usepackage{pifont}% http://ctan.org/pkg/pifont
\usepackage{hyperref}
\usepackage[nameinlink]{cleveref}
\usepackage{soul}
\usepackage{algorithm}
\usepackage{algpseudocode}
\usepackage{booktabs}
\usepackage{multirow}
\usepackage{arydshln}
\usepackage{adjustbox}
\usepackage{lipsum}
\usepackage{xcolor}
\usepackage{wrapfig}
\usepackage{subcaption}
\usepackage{graphicx}
\usepackage{bbm}

\definecolor{light-gray}{gray}{0.9}
\definecolor{light-blue}{RGB}{100, 100, 255}

\newcommand{\react}[0]{\textsc{ReACT}}

\newcommand{\dynathink}[0]{Dyna-Think}
\newcommand{\framework}[0]{Dyna-Mind}
\newcommand{\DFSfull}[0]{{Reasoning with Simulations}}
\newcommand{\DFS}[0]{\textsc{ReSim}}
\newcommand{\DFSsymbol}[0]{ReSim}
\newcommand{\Distill}[0]{\textsc{Distill}}

\newcommand{\wmrolloutshort}[0]{\textsc{SimRollout}}
\newcommand{\RLWM}[0]{{Dyna-GRPO}}
\newcommand{\RLWMshort}[0]{\textsc{Dyna-GRPO}}

\title{\framework{}: Learning to Simulate from\\ Experience for Better AI Agents}

\author{Xiao Yu$^1$\thanks{Work done during internship at Microsoft Research; $^\dagger$Project Lead}\,\,, \, Baolin Peng$^{2\dagger}$, \, Michel Galley$^2$, \, Hao Cheng$^2$, \, Qianhui Wu$^2$\\ \bf{Janardhan Kulkarni$^{2}$,} \,\bf{Suman Nath$^{2}$,} \,\bf{Zhou Yu$^{1}$,} \,\bf{Jianfeng Gao$^{2}$} \\
$^1$Columbia University, NY \, $^2$Microsoft Research, Redmond \\
\texttt{\{xy2437, zy2461\}@columbia.edu} \\
\texttt{\{baolinpeng, jfgao\}@microsoft.com}
}

\begin{document}

\maketitle

\begin{abstract}
    Reasoning models have recently shown remarkable progress in domains such as math and coding.
    However, their expert-level abilities in math and coding contrast sharply with their performance in long-horizon, interactive tasks such as web navigation and computer/phone-use.
    Inspired by literature on human cognition, we argue that current AI agents need ``vicarious trial and error''---the capacity to mentally simulate alternative futures before acting---in order to enhance their understanding and performance in complex interactive environments.
    We introduce \framework{}, a two-stage training framework that explicitly teaches (V)LM agents to integrate such simulation into their reasoning.
    In stage 1, we introduce \DFSfull{} (\DFS{}), which trains the agent to generate structured reasoning traces from expanded search trees built from real experience gathered through environment interactions.
    \DFS{} thus grounds the agent's reasoning in faithful world dynamics and equips it with the ability to anticipate future states in its reasoning.
    In stage 2, we propose \RLWM{}, an online reinforcement learning method to further strengthen the agent's simulation and decision-making ability by using both outcome rewards and intermediate states as feedback from real rollouts. Experiments on two synthetic benchmarks (Sokoban and ALFWorld) and one realistic benchmark (AndroidWorld) demonstrate that (1) \DFS{} effectively infuses simulation ability into AI agents, and (2) \RLWM{} leverages outcome and interaction-level signals to learn better policies for long-horizon, planning-intensive tasks.
    Together, these results highlight the central role of simulation in enabling AI agents to reason, plan, and act more effectively in the ever more challenging environments.
\end{abstract}

\section{Introduction}

%% suggested v2 of intro:

Recent advances in language models have unlocked impressive reasoning capabilities in domains such as mathematics and programming \citep{shao2024deepseekmathpushinglimitsmathematical,jimenez2024swebenchlanguagemodelsresolve}.
However, many emerging applications unfold in complex environments that require multi-step reasoning, such as web navigation \citep{zhou2024webarenarealisticwebenvironment,deng2023mind2webgeneralistagentweb}, deep research \citep{gou2025mind2web2evaluatingagentic,du2025deepresearchbenchcomprehensivebenchmark}, and computer/phone-use tasks \citep{xie2024osworldbenchmarkingmultimodalagents,rawles2025androidworlddynamicbenchmarkingenvironment}. Success in these domains depends not only on the ability to decompose goals and reflect on past progress, but also on AI agents' ability to construct accurate world models that capture the structure and dynamics of increasingly complex environments
\citep{shao2024deepseekmathpushinglimitsmathematical,jimenez2024swebenchlanguagemodelsresolve}.

Insights from human cognition indicate why such ability to model and simulate complex environments is critical. Neuroscience research \citep{tolman1948cognitive,Daw2005UncertaintybasedCB,daw2014algorithmic,bennett2023brief} highlights the emergence of the neocortex as a turning point in intelligence, enabling early mammals to engage in ``vicarious trial and error'': mentally simulating possible futures, evaluating their consequences, and selecting advantageous actions without directly experiencing each option. This ability greatly enhanced adaptability and decision-making, which we argue is equally essential for reasoning in long-horizon AI agent tasks.

% \begin{wrapfigure}[15]{L}{0.5\textwidth}
%     \vspace{-10mm}
%     % \begin{figure}[h]
%         \centering
%         \includegraphics[scale=0.8]{./images/overall_pipe_coverfig_cropped.pdf}
%         \vspace{-3mm}
%         \caption{\framework{}}
%         \label{fig:overall_pipeline_fig}
%     % \end{figure}
% \end{wrapfigure}
% \begin{wrapfigure}[12]{R}{0.5\textwidth}
%     \vspace{-10mm}
%     % \begin{figure}[h]
%         \centering
%         \includegraphics[scale=0.4]{./images/r1_wm_plot.pdf}
%         \vspace{-3mm}
%         \caption{DeepSeek-R1 show strong world simulation ability in Sokoban, but not in ALFWorld.}
%         \label{fig:r1_wm_fig}
%     % \end{figure}
% \end{wrapfigure}
\begin{figure}[ht]
    \centering
    \begin{subfigure}[t]{0.38\textwidth}
        \centering
        \includegraphics[scale=0.28]{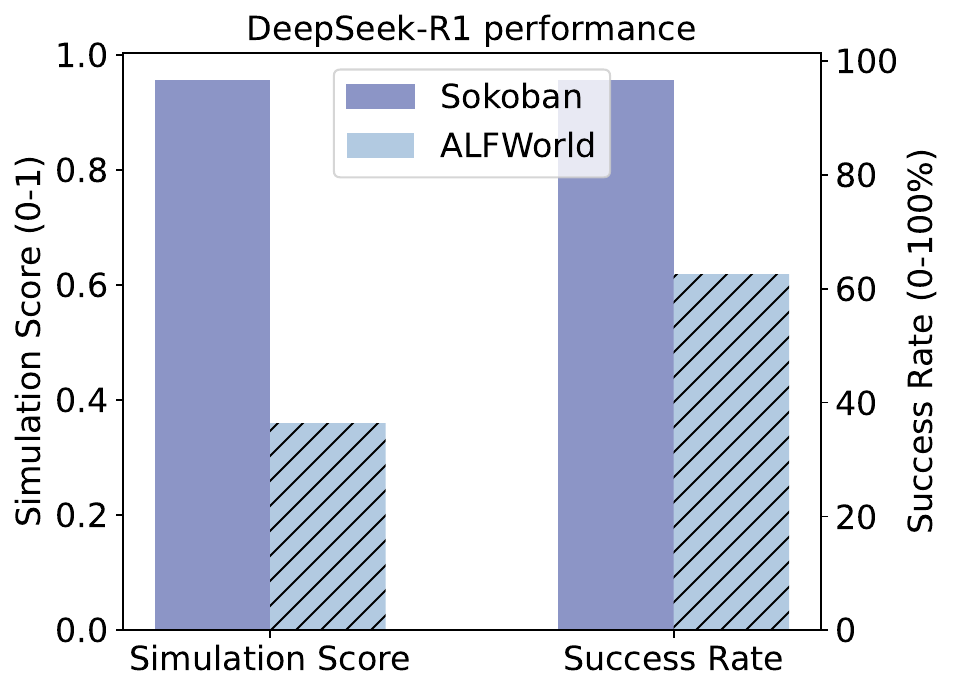}
        \caption{Simulation ability v.s. performance}
        \label{fig:r1_wm_fig}
    \end{subfigure}
    \hfill
    \begin{subfigure}[t]{0.61\textwidth}
        \centering
        \includegraphics[scale=0.82]{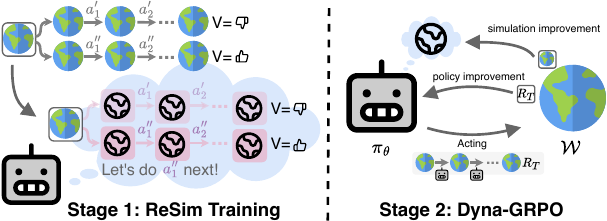}
        \caption{\framework{}}
        \label{fig:overall_pipeline_fig}
    \end{subfigure}
    \caption{We find the performance of strong reasoning models is heavily affected by its ability to simulate in different environments (\textbf{left}). We introduce \framework{}, a two-stage training framework to integrate and improve simulation ability of AI agents (\textbf{right}).}
    % \vspace{-10mm}
\end{figure}
% Empirical evidence supports this view. In \Cref{fig:r1_wm_fig}, we find while strong reasoning models such as DeepSeek-R1 can successfully simulate and solve structured environments like Sokoban, their performance drops sharply in more complex domains such as ALFWorld - both in terms of simulation accuracy and overall task success.
% While there have been initial attempts to mitigate this limitation, such as Dyna-Think \citep{yu2025dynathinksynergizingreasoningacting} which integrates simulation ability into reasoning by distilling simplified traces and adding auxiliary next-state prediction tasks, these approaches rely on world simulation data generated by reasoning models themselves, which may encode errors or biases and thus limit reliability.
Empirical evidence supports this view. In \Cref{fig:r1_wm_fig}, we observe that while strong reasoning models such as DeepSeek-R1 can simulate and solve structured environments like Sokoban, their performance drops sharply in more complex domains such as ALFWorld---both in simulation accuracy and overall task success (also see \Cref{subsubsec:World Model Accuracy}).
Initial attempts to address this limitation, such as Dyna-Think \citep{yu2025dynathinksynergizingreasoningacting}, integrate simulation into reasoning through distilling simplified traces and adding auxiliary next-state prediction tasks.
% However, these methods rely on distilling (simplified) simulation data \emph{generated} by the reasoning models themselves, which can embed errors or biases and limit reliability.
% However, these methods rely on learning from synthetic simulation data generated directly by reasoning models themselves, which can embed errors or biases and limit reliability.
However, these methods rely on the strong capability of reasoning models to directly generate synthetic simulation data, which can embed errors and biases.

% To overcome this limitation, we present \framework{}...
% This limitation highlights a key hypothesis: for robust simulation, it is not sufficient to rely on reasoning traces that may encode errors or biases. Instead, learning directly from real environment interactions provides more faithful and grounded supervision for developing world models, by which agents can acquire simulation capinabilities that are both more accurate and more aligned with the environments they act in.
To overcome this limitation, we present \framework{}, an improved two-stage training framework to teach (V)LM agents to simulate the environment by directly learning from real experiences.
In stage 1 training, we propose \DFSfull{} (\DFS{}) to algorithmically construct reasoning traces using expanded search trees obtained from real environment interactions, and train a policy model using these reasoning traces.
In stage 2 training, we further improve the policy and its simulation ability using online reinforcement learning (RL).
We introduce \RLWM{}, a novel algorithm that utilizes both outcome rewards and intermediate states from rollouts to improve the simulation ability of the policy.
Extensive experiments on two widely used synthetic benchmarks (Sokoban and ALFWorld) and one realistic benchmark (AndroidWorld) show the effectiveness of each stage of the framework.
Our results indicate that 
% 1) reasoning traces constructed by \DFS{}, while being 7x shorter, is more effective than that from existing models such as DeepSeek-R1; and 2) both training stages of \framework{} successfully improved both the model's world modeling ability and task performance compared to baselines.
(1) \DFS{}'s reasoning traces effectively teach AI agents to simulate;
and (2) \RLWM{}, by leveraging both outcome rewards and intermediate interactions, learns better policies for long-horizon, planning-intensive tasks.
These findings highlight the importance of world simulation ability for reasoning in long-horizon tasks.

\section{Related Work}
\label{sec:Related Work}

\paragraph{(V)LM as decision making agents}
The use of (visual) language models as autonomous agents has been explored in a wide range of applications such as interactive game playing \citep{wang2023voyageropenendedembodiedagent,feng2025groupingrouppolicyoptimizationllm}, computer, phone, and browser uses \citep{xie2024osworldbenchmarkingmultimodalagents,zhou2024webarenarealisticwebenvironment,rawles2025androidworlddynamicbenchmarkingenvironment}, software engineering \citep{jimenez2024swebenchlanguagemodelsresolve,yang2024sweagentagentcomputerinterfacesenable}, and more.
Early works include reactive agents \citep{yao2023reactsynergizingreasoningacting} that directly prompts an (V)LM to make decisions on immediate observations without simulation or planning approaches, hindering performance on complex long-horizon tasks.
Recent advances include: (1) search-based methods \citep{yao2023treethoughtsdeliberateproblem,zhou2024languageagenttreesearch,koh2024treesearchlanguagemodel, yu2023promptbasedmontecarlotreesearch,yu2025exactteachingaiagents} that augments (V)LM agents with algorithms such as BFS, DFS, and MCTS; and (2) hierarchical, multi-agent methods \citep{zheng2024gpt4visiongeneralistwebagent,agashe2024agentsopenagentic,agashe2025agents2compositionalgeneralistspecialist,liu2025pcagenthierarchicalmultiagentcollaboration,gou2025navigatingdigitalworldhumans} that orchestrate multiple specialized agents to complete long-horizon tasks.
While these methods show improvements, they often introduce substantial overheads during inference, such as requiring additional interactions with the environments or designing complex heuristics to orchestrate multiple agents.
% In this work, we focus on \emph{training} a single (V)LM agent capable of planning via simulation during its reasoning to enhance its decision making ability.
We focus on enhancing a single (V)LM agent by integrating simulation into its reasoning via training.

\paragraph{Training (V)LM agents}
Early methods in training (V)LM agents mostly rely on supervised learning (SFT) with human annotations or data synthesized by state-of-the-art (reasoning) models \citep{zeng2023agenttuningenablinggeneralizedagent,chen2024agentflandesigningdatamethods,zhang2024xlamfamilylargeaction,xu2025aguvisunifiedpurevision}.
Recently, many methods such as \citet{feng2025groupingrouppolicyoptimizationllm,wang2025ragenunderstandingselfevolutionllm,wei2025swerladvancingllmreasoning,wei2025webagentr1trainingwebagents} leverage reinforcement learning (RL) with verifiable rewards to directly train agents to complete tasks by \emph{prompting} them to reason before taking actions, following the success of DeepSeek-R1 \citep{deepseekai2025deepseekr1incentivizingreasoningcapability}.
However, it remains unclear whether extensive reasoning is necessary for all scenarios \citep{shojaee2025illusionthinkingunderstandingstrengths}, and what aspects of such reasoning is essential for long-horizon tasks \citep{yu2025dynathinksynergizingreasoningacting}.
In this work, we specialize in integrating and improving the simulation ability of (V)LM agents during reasoning, and show that planning with world simulation is crucial for long-horizon tasks.
% To more effectively advance model capability, we construct our reasoning data using non-reasoning models such as DeepSeek-V3 \citep{deepseekai2025deepseekv3technicalreport}.

\paragraph{World models and Dyna algorithms}
Beyond task completion, real-world interaction data contains rich information that can be used to help decision making.
Early examples include Dyna algorithms \citep{dyna}, which combine model-based and model-free methods to efficiently learn optimal policies. Given a set of real-world rollout data, Dyna (1) separately train a world model using these rollouts; (2) perform additional simulated rollouts with the world model; and (3) update the policy using both real and simulated rollouts.
Applications of world model training have been explored in work such as \citet{chae2025webagentsworldmodels,gu2025llmsecretlyworldmodel}, facilitating search algorithms such as MCTS to improve performance; and applications of Dyna include Deep Dyna-Q \citep{peng2018deepdynaqintegratingplanning}, Switch-DDQ \citep{wu2018switchbasedactivedeepdynaq}, and more \citep{pseudo-dyna-q,yu2025dynathinksynergizingreasoningacting}.
However, these approaches either result in modular systems (a separate policy and world model) or require accessing state-of-the-art reasoning models (e.g., DeepSeek-R1).
% Our work outperforms strong reasoning models such as DeepSeek-R1 while only using weaker, non-reasoning models (e.g., DeepSeek-V3).
Our work does not rely on strong reasoning models, and focuses on integrating and improving simulation as part of an agent's reasoning process.

\vspace{-1mm}
\section{\framework{}}
\label{sec:Methods}
% Research on human problem-solving suggests that a key component that enables long-horizon planning is the ability to \emph{reason} with an internal world model.

Research in human cognition \citep{Daw2005UncertaintybasedCB,daw2014algorithmic,bennett2023brief} as well as in games like chess, go, and othello \citep{muzero,li2024emergentworldrepresentationsexploring, nanda2023emergentlinearrepresentationsworld,chae2025webagentsworldmodels} suggests that strong agents implicitly store and use a (compressed) representation of the world to enhance their decision-making.
This perspective highlights two key questions in existing approaches to improve (V)LM agents for long-horizon tasks: (1) how to synergize world simulations with reasoning; and (2) how to improve the simulation ability to help improve the policy.

To address these questions, we introduce \framework{}, a two-stage training framework to teach (V)LM agents to plan with simulations during their reasoning and improve their task performance.
We detail these two training stages next in \Cref{sec:DFS Distillation} and \Cref{subsec:Online RL}, respectively.
% To initialize the ability to reason with an internal world model, we propose \DFS{} Distillation...

% While we find such method significantly improves both the world modeling ability and policy performance, this method is computationally expensive as it relies on multiple models to synthesize such data.
% To this end, we introduce \RLWMshort{}, a simple modification on top of GRPO to further improve the agent's policy and world model ability during online RL training.

% In the remainder of this section, we detail these two training stages in \Cref{sec:DFS Distillation} and \Cref{subsec:Online RL}, respectively.

\subsection{Notation}
\label{subsec:Problem Setup}
Completing tasks in complex, realistic environments is typically formulated as a Markov Decision Process of $(\mathcal{S}, \mathcal{A}, \mathcal{T}, \mathcal{R})$.
In the generic setting of multi-step tasks, an agent $\pi_\theta$ receives an instruction and observation\footnote{
Technically, any input to the agent from our environments is an observation (as in POMDP) instead of a state. However, to simplify notation we used $s$ to generally denote the agent's input from the environment.
} from the environment $s_{t} \sim \mathcal{S}$ at time step $t$, generates an action $a_{t} \sim \pi_\theta(\cdot|s_t)$, and transitions to the next state $s_{t+1} \sim \mathcal{T}(s_t, a_t)$.
This process is repeated until the task is completed or until reaching a maximum number of steps, upon which a terminal reward $r_{T} \sim \mathcal{R}(s_T, a_T)$ is provided based on whether the task is completed successfully or not.
In the context of simple text games such as Sokoban \cite{SchraderSokoban2018}, a state $s_t$ can represent the complete game state, and an action $a_t$ is one of ``left'', ``right'', ``up'', ``down'' (after some reasoning process).
In more complex environments such as AndroidWorld \citep{rawles2025androidworlddynamicbenchmarkingenvironment}, a state $s_t$ is the current screenshot of the android device, and an action $a_t$ can be ``tapping on a coordinate (x,y)'', ``swiping up'', ``swiping down'', etc.
We note that since we aim to train agents to generate simulations \emph{within} their reasoning process, any text that represents simulation is always part of the response $a_t$.\footnote{
As action plan/final action are always extracted from model response, we use $a$ (by slight abuse of notation) to denote either the full response or the extracted executable action. Distinctions are made clear in context. Example model response for each benchmark is provided in \Cref{tab:sokoban_example}, \Cref{tab:alfworld_example}, and \Cref{fig:android_example}.
} Any variant of the symbol $s$ represents \emph{real} states from environment interactions, unless explicitly stated otherwise.

\begin{figure}[t!]
    \centering
    \includegraphics[scale=0.85]{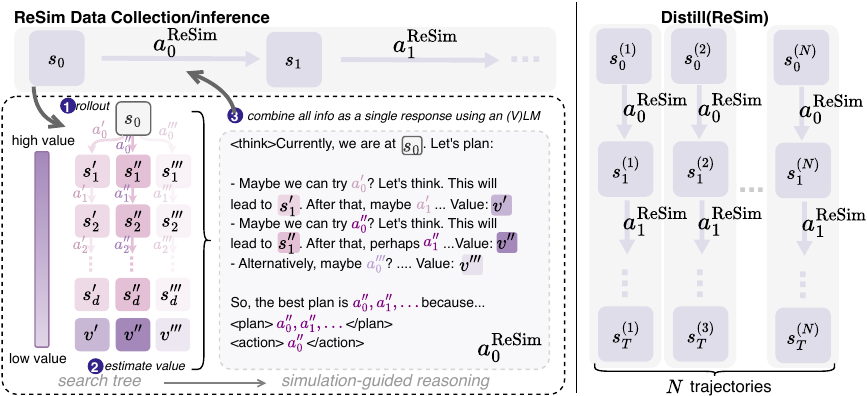}
    \caption{
    % Overview of \DFS{} for data collection (left) and for distillation training (right) during stage-1 training. Given a state $s$, \DFS{} first uses a rollout model $\pi_\theta$ to generate $b$ rollouts upto depth $d$, and then uses $V_\nu$ to estimate the value of each rollout. Given $b$ rollouts and their values, \DFS{} finally uses a (V)LM to aggregate all information into a single response $a^{\mathrm{\DFSsymbol}}$ for the current state $s$. The resuling trajectory $(s_0, a_1^{\mathrm{\DFSsymbol}}, s_1, a_2^{\mathrm{\DFSsymbol}}, ...)$ is then used for SFT distillation training.
    \DFS{} integrates simulation into reasoning ($a_t^{\mathrm{\DFSsymbol}}$) by using expanded search trees built through \emph{real} environment interactions (\textbf{left}).
    \DFS{} then trains an agent to directly generate such simulation-guided reasoning trace  $a_t^{\mathrm{\DFSsymbol}}$ without any algorithm support (\textbf{right}).
    }
    \label{fig:dfs_algo_fig}
    \vspace{-4mm}
\end{figure}
\subsection{\DFSfull{} (\DFS{})}
\label{sec:DFS Distillation}

To enable an agent to simulate during its reasoning, we first construct imitation learning data where the reasoning process consists of explicitly planning with simulations.
% Different from prior work such as \citet{yu2025dynathinksynergizingreasoningacting} that leverages superior LLMs such as DeepSeek-R1 \citep{deepseekai2025deepseekr1incentivizingreasoningcapability} which already shows world modeling capability in its reasoning traces (see \Cref{subsubsec:Toy Text Main Results} for more details), we construct rich simulation-based reasoning traces using \emph{real environment interactions} summarized by a weaker, non-reasoning LLM such as DeepSeek-V3 \citep{deepseekai2025deepseekv3technicalreport}.
Different from prior work such as \citet{yu2025dynathinksynergizingreasoningacting} that leverages superior LLMs such as DeepSeek-R1 which already shows world modeling capability in its reasoning traces (see \Cref{subsubsec:Toy Text Main Results} for more details), we construct simulation-guided reasoning traces using search trees built from \emph{real environment interactions}.

\paragraph{\DFS{} Data Collection} To construct reasoning data with rich simulations, we leverage algorithms such as depth first search (DFS) to construct search trees based on environment interactions, and then use an (V)LM to aggregate \emph{the entire search tree} into a single reasoning response $a^{\mathrm{\DFSsymbol}}$ for later training. Specifically, given a state $s$, \DFS{} first uses a rollout model $\pi_\theta$ to generate $b$ rollouts from $s$ up to depth $d$. This rollout model can be a specialized/finetuned LLM (see \Cref{subsec:Toy Text}) or simply prompting a generic LLM (see \Cref{subsec:AndroidWorld}).
Then, \DFS{} uses a value function $V_\nu$ to provide an estimate of the quality of each of the partial rollouts, where the $V_\nu$ can be implemented as either a finetuned value model (see \Cref{subsec:Toy Text}) or using LLM-as-a-judge (see \Cref{subsec:AndroidWorld}).
Finally, we use a generic (V)LM to aggregate all these rollouts and their values into a single response $a^{\mathrm{\DFSsymbol}}$ by prompting the (V)LM to 1) first independently summarize each partial rollout, \emph{which contains ground-truth future states information from the environment}; and 2) then aggregate all these summaries into a coherent response conditioned on the current state $s$ and previous $h$ actions and states, and choose the best plan and the next immediate action for execution.
The final chosen action from $a^{\mathrm{\DFSsymbol}}$ is then executed in the environment, and this process is repeated until the task is solved or until a maximum number of steps is reached.
We illustrate this process in \Cref{fig:dfs_algo_fig} Left and \Cref{algo:ourdfs}.
% We note that since \DFS{} essentially converts real search trees into a single reasoning trace, other algorithms such as A* search or MCTS could also be used, which we leave for future work.
We note that since \DFS{} essentially converts real search trees into a single reasoning trace, it is not limited to (1) agent-environment interactions; (2) specific search algorithms used in this work. We believe other domains such as agent-user-environment interactions or other algorithms such as MCTS are also applicable, which we leave for future work.

% \paragraph{\DFS{} Distillation} Since each action $a^{\mathrm{\DFSsymbol}}$ encapsulates an entire search tree in its reasoning by construction, we directly use $a^{\mathrm{\DFSsymbol}}$ as the training target to teach a model to mimic the ``search'' process and hence synergize simulation with reasoning.
\paragraph{\DFS{} Distillation} Since each response $a^{\mathrm{\DFSsymbol}}$ encapsulates an entire search tree in its reasoning, we directly use $a^{\mathrm{\DFSsymbol}}$ as the training target given an input $s$ to teach the model to perform simulation-guided reasoning without any algorithm support.
We illustrate this in \Cref{fig:dfs_algo_fig} Right.
Specifically, given a collection of trajectories $\tau = \{s_0, a^{\mathrm{\DFSsymbol}}_0, s_1, a^{\mathrm{\DFSsymbol}}_1, \cdots, s_{T}, a^{\mathrm{\DFSsymbol}}_T\}$ produced by \DFS{} inference, we use SFT to train the model to directly generate each $a_t^{\mathrm{\DFSsymbol}}$ given the current state $s_{t}$ as well as a maximum history of $h$ previous actions and states in the trajectory (i.e., the same input used by other inference methods such as \react).

% \subsection{\RLWMshort{} Training}
\subsection{\RLWM{}}
\label{subsec:Online RL}

While \DFS{} provides a principled way to synergize simulation with reasoning, it is computationally expensive and relies on multiple modules (a rollout model, a value function, and a (V)LM to aggregate the search tree into a single response) to construct training data.
Additionally, such offline training may limit models' generalization ability to new tasks.
To address this, we propose \RLWMshort{}, a modification of GRPO \citep{shao2024deepseekmathpushinglimitsmathematical} to further improve the model's simulation ability during online RL without using any search or additional modules.
% \todo{} we may also need to talk about episode-relative advantage?
% To further improve the models simulation ability, we use online RL. However, world model signals such as $s' \sim T(s, a)$ contains \emph{rich text/visual information}, whereas conventional RL algorithms such as PPO or GRPO aims to learn a policy based solely on \emph{scalar} rewards. In this work, we propose \RLWMshort{}, a simple modification on top of GRPO to incorporate world model signals into the training process. 
The standard GRPO objective $\mathcal{J}_{\textrm{GRPO}}$ is:
% \begin{align}
% &\mathcal{J}_{\textrm{GRPO}} = \\
%  &\,\,\mathbb{E}_{\tau \sim \pi_{\theta_{\mathrm{old}}}} \left[ 
%     \frac{1}{GT} \sum_{i=1}^{G} \sum_{t=1}^{T} 
%     \min{ \left(\rho_\theta(a^{(i)}_t)A(a^{(i)}_t), \textrm{clip}(\rho_\theta(a^{(i)}_t), 1\pm\epsilon) A(a^{(i)}_t) \right)}  -\beta D_{\textrm{KL}}(\pi_\theta || \pi_{\theta_{\mathrm{ref}}}) \right],\nonumber 
% \end{align}
\begin{align}
    \mathbb{E}_{\tau \sim \pi_{\theta_{\mathrm{old}}}} \left[ 
    \frac{1}{GT} \sum_{i=1}^{G} \sum_{t=1}^{T} 
    \min{ \left(\rho_\theta(a^{(i)}_t)A(a^{(i)}_t), \textrm{clip}(\rho_\theta(a^{(i)}_t), 1\pm\epsilon) A(a^{(i)}_t) \right)}  -\beta D_{\textrm{KL}}(\pi_\theta || \pi_{\theta_{\mathrm{ref}}}) \right],\nonumber 
\end{align}
where $\rho_\theta(a) = \frac{\pi_\theta(a|s)}{\pi_{\theta_{\mathrm{ref}}}(a|s)}$ is the importance sampling ratio, $\beta$ is the KL regularization coefficient, and $A=A_{\mathrm{GRPO}}$ is the episode-level advantage function \citep{wang2025ragenunderstandingselfevolutionllm,feng2025groupingrouppolicyoptimizationllm}:
\[
    A(a_t^{(i)})=A_{\mathrm{GRPO}}(\tau^{(i)}) = \frac{R(\tau^{(i)}) - \textrm{mean}(\{R(\tau^{(j)})\}_{j=1}^{G})}{\textrm{std}(\{R(\tau^{(j)})\}_{j=1}^{G})},\quad R(\tau^{(i)}) = \sum_{t=1}^{T} R(s_t, a_t),
\] 
where $G$ is the group size, $R(\cdot)$ is the reward provided by the environment, with $R(s_t, a_t)=-0.1$ for non-terminal steps and $R(s_T, a_T)=10.0$ or $R(s_T, a_T)=0.0$ for terminal steps when task succeeded or failed, respectively.

% However, RL algorithms such as GRPO solely focus on improving a policy using \emph{scalar} rewards such as task success but does not provide any direct training signal on refining the reasoning process or world model simulations.
% We propose \RLWMshort{} to address this, by additionally incorporating future state(s) information $s_{t+1}, s_{t+2}, \cdots$ as \emph{textual} signals to help improve model's response $a \sim \pi_\theta(\cdot |s_t)$ at state $s_t$ during RL training.
However, RL algorithms such as GRPO aim to optimize a policy only using \emph{scalar} rewards $R_T$ but do not provide any direct training signal on refining the reasoning process or world model simulations.
We propose \RLWMshort{} to address this, by \emph{additionally} incorporating future state(s) information $s_{t+1}, s_{t+2}, \cdots$ as \emph{textual} signals to help improve the model's response $a \sim \pi_\theta(\cdot |s_t)$ during RL training.
% To achieve this, we utilize insights from prior work such as Reflexion \citep{shinn2023reflexionlanguageagentsverbal} to directly prompt the underlying model to refine its simulation in $a \sim \pi_\theta(\cdot |s_t)$ given real future state(s) $s_{t+1}, s_{t+2}, \cdots$ during RL rollouts (\wmrolloutshort{}).
% Then, during optimization we train the policy to both directly generate the refined action and to improve its ``simulation refinement'' ability (\RLWMshort{}). We detail these two modifications below.
Since textual signals cannot be directly ``optimized'', we propose \wmrolloutshort{} to instead \emph{prompt the underlying model} to refine its simulation in $a \sim \pi_\theta(\cdot |s_t)$ utilizing real future state(s) $s_{t+1}, s_{t+2}, \cdots$ \emph{during RL rollouts}.
Then, during optimization we train the policy to both directly generate the refined action and also to improve its ``simulation refinement'' ability (\RLWMshort{}). We detail these two modifications below.

\begin{figure}[t!]
    \centering
    \includegraphics[scale=0.63]{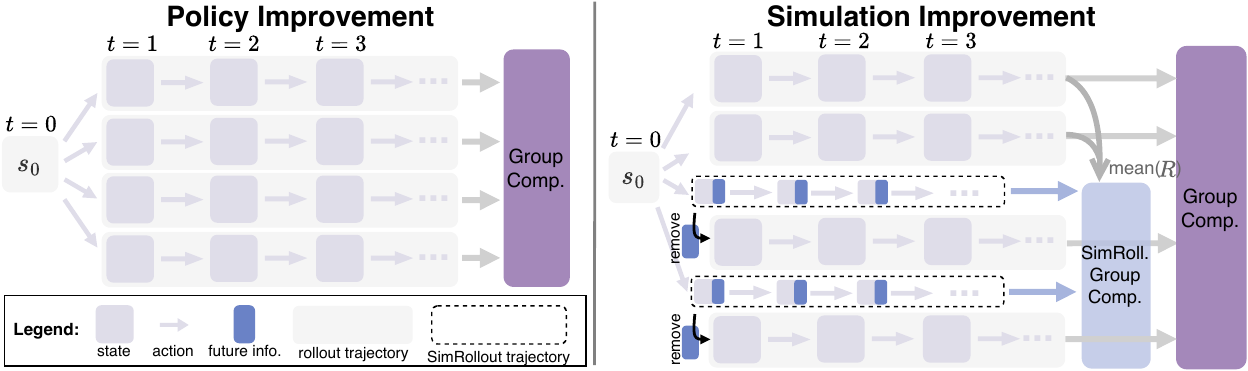}
    \caption{\RLWMshort{} iterates between policy improvement (\textbf{left}) and world model improvement (\textbf{right}), optimized by GRPO.
    During policy improvement, we perform grouped policy rollouts with GRPO.
    During simulation improvement, we perform both policy rollouts and simulation refinement rollouts (see \Cref{fig:selfimp_rollout_fig}), and trains the model to directly \textcolor{violet}{generate an improved policy} as well as to \textcolor{light-blue}{better perform simulation refinement} when provided with future-states information.}
    \label{fig:rl_algo_fig}
    \vspace{-4mm}
\end{figure}
\begin{wrapfigure}[12]{R}{0.48\textwidth}
    \vspace{-11mm}
    % \begin{figure}[h]
        \centering
        \includegraphics[scale=1.0]{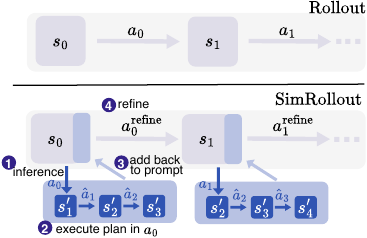}
        \caption{\wmrolloutshort{} generates refined action per state $s_t$ using real environment interactions}
        \label{fig:selfimp_rollout_fig}
    % \end{figure}
\end{wrapfigure}
\paragraph{\wmrolloutshort{}} 
In \underline{sim}ulation \underline{r}efinement r\underline{ollout} (\wmrolloutshort{}), at each state $s_t$ we first sample a response $a \sim \pi_\theta(\cdot|s_t)$; then extract the final chosen plan $\{\hat{a}_1, \hat{a}_2, \cdots, \hat{a}_d\}$ up to depth $d$ from $a$ and execute them in the environment to obtain ground truth next-states $\{s'_{t+1}, s'_{t+2}, \cdots, s'_{t+d}\}$; and finally prompt $\pi_\theta$ again to refine its response $a$ given these real future states $a^{\textrm{refine}} \sim \pi_\theta(\cdot | s^{\textrm{refine}}_t) , s^{\textrm{refine}}_t \equiv \{s_{t} \oplus a \oplus s'_{t+1} \oplus \hat{a}_2 \oplus \cdots \oplus s'_{t+d}\}$.
We illustrate this rollout process in \Cref{fig:selfimp_rollout_fig} and provide the pseudo-code in \Cref{algo:wmimp}. 
We note that this is different from methods such as Reflexion \citep{shinn2023reflexionlanguageagentsverbal}, which performs reflection at the end of the episode utilizing success/failure information, and is also not intended for any training purposes.
Empirically, we find the resulting $a^{\textrm{refine}}$ indeed improves the policy's simulation and performance (see \Cref{subsec:World Model Refinement Performance}).
% Specifically, during such world model training, we generate improved trajectories $\tau_{\mathrm{imp}}$ by 1) first sampling an response from $a \sim \pi_\theta(a|s_t)$, 2) then extract the action plan $\{a_1, a_2, \cdots, a_t\}$ from $a$ and executing them in the environment to obtain ground truth next-states $\{s_{t+1}, s_{t+2}, \cdots, s_{t+t}\}$; and 3) finally prompting the model to refine its action $a$ provided with these next-states information (see \Cref{algo:wmimp}).

\begin{wrapfigure}[16]{R}{0.5\textwidth}
    \vspace{-4mm}
\scalebox{0.8}{
    \begin{minipage}{0.6\textwidth}
        \begin{algorithm}[H]
            \caption{\RLWMshort{}}\label{algo:rlwm}
            \begin{algorithmic}[1]
            % \Repeat
            \Require policy $\pi_\theta$, environment $\mathcal{T}$, group size $G$
            \Require hyperparameters $G, N, n_{\mathcal{T}}, n_{\pi}$
            \For {$N$ training iterations}
                \State \emph{\textcolor{gray}{// simulation improvement}}
                \For {$n_{\mathcal{T}}$ steps}
                    \State \emph{\textcolor{gray}{// see \Cref{algo:wmimp}}}
                    \State $\{\tau'\}, \{\tau'_{\textrm{refine}}\} \gets$ SimRollout$(\pi_\theta, \mathcal{T}, G/2)$
                    \State $\{\tau\} \gets$ Rollout$(\pi_\theta, \mathcal{T}, G/2)$
                    \State Update $\pi_\theta$ with GRPO$(\{\tau\} \cup \{\tau'\})$
                    \State Update $\pi_\theta$ with GRPO$(\{\tau'_{\textrm{refine}}\})$ using $A_{\mathrm{refine}}$
                \EndFor
                \State \emph{\textcolor{gray}{// policy improvement}}
                \For {$n_{\pi}$ steps}
                    \State $\{\tau\} \gets$ Rollout$(\pi_\theta, \mathcal{T}, G)$
                    \State Update $\pi_\theta$ with GRPO$(\{\tau\})$
                \EndFor
            \EndFor
            \State \Return $\pi_\theta$
            \end{algorithmic}
        \end{algorithm}
    \end{minipage}
}
\end{wrapfigure}
% \paragraph{World Model Improvement}
\paragraph{\RLWMshort{} Training}
To utilize refined trajectories from \wmrolloutshort{} during RL, we follow Dyna algorithms to improve the model's policy and simulation ability iteratively.
% To use these improved actions during online RL training, we follow Dyna algorithms and iterate between 1) world model improvement where models are trained on improved policies using next-state information; and 2) policy improvement where models are trained on normal rollouts from the policy.
% Specifically, \RLWMshort{} iterates between 1) simulation improvement where models are trained on refined policies that uses future states information from \wmrolloutshort{} to improve its simulation ability; and 2) direct policy improvement where models are trained with normal rollouts without accessing future states to better integrate its simulation ability into decision-making.
Specifically, \RLWMshort{} iterates between (1) \emph{simulation improvement} where models learn from refined policies that use future states information from \wmrolloutshort{} to improve its simulation ability; and (2) direct \emph{policy improvement} where models are trained on standard rollouts without future-state access, allowing it to better integrate simulation ability into decision-making.
We illustrate both training processes in \Cref{fig:rl_algo_fig}, and detail the overall algorithm in \Cref{algo:rlwm}.

% During world model improvement, we train $\pi_\theta$ to directly generate the improved actions $a^{imp}$ given the current state $s_t$, as well as to better perform ``self improvement'' when prompted with next-state information.
During simulation improvement, for each task we (1) first perform \wmrolloutshort{} with a group size of $G/2$, collecting refined trajectories with and without future-state information removed: $\tau' = \{s_0, a^{\textrm{refine}}_0, s_1, a^{\textrm{refine}}_1, \cdots\}$ and $\tau'_{\mathrm{refine}} = \{s_0^{\textrm{refine}}, a_0^{\textrm{refine}}, s_1^{\textrm{refine}}, a_1^{\textrm{refine}}, \cdots\}$; (2) then perform standard rollouts with group size of $G/2$;
(3) combine these standard rollouts $\tau$ with refined trajectories $\tau'$ into a single group of size $G$ and perform GRPO on this combined group; 
% 3) finally, to train the model to improve its ability to perform world model refinement, we also utilize $\tau'_{\mathrm{refine}}$ as training data and modify the advantage term in GRPO to incentivize successful refinements that both correctly solves the task \textbf{and} improves upon (the mean reward of) the normal policy:
(4) finally utilize $\tau'_{\mathrm{refine}}$ to also improve the model's simulation refinement ability, using the following modified advantage to reward refinements that both correctly solves the task \emph{and} improves upon (the mean reward of) standard policy rollouts which does not access future states:
% To achieve the former, we perform GRPO treating these improved trajectories $\tau' = \{s_0, a^{\textrm{imp}}_1, s_1, a^{\textrm{imp}}_2, \cdots\}$ as if they are normal rollouts $\tau = \{s_0, a_1, s_1, a_2, \cdots\}$. To achieve the latter, we modify advantage term in GRPO to a binary reward to learn successful self-improvements that both correctly solves the task \textbf{and} improves upon (the mean reward of) the normal policy:
\[
    A_{\mathrm{refine}}(\tau_{\mathrm{refine}}^{(i)}) = \begin{cases}
        1.0, & \text{if } \tau_{\mathrm{refine}}^{(i)} \text{ is correct and } R(\tau_{\mathrm{refine}}^{(i)}) > \max(\bar{R}, \bar{R}^{\textrm{refine}}) \\
        0.0, & \text{otherwise}
    \end{cases},
\] 
where $\bar{R} = \frac{1}{G/2} \sum_{i=1}^{G/2} R(\tau^{(i)})$ is the mean reward of the standard policy rollouts (line 6 of \Cref{algo:rlwm}); $\bar{R}^{\textrm{refine}} = \frac{1}{G/2} \sum_{i=1}^{G/2} R(\tau_{\mathrm{refine}}^{(i)})$ is mean reward from \wmrolloutshort{} (line 5 of \Cref{algo:rlwm}). During policy improvement, we perform standard policy rollouts without future state information, optimized by GRPO using episode-level advantage \citep{feng2025groupingrouppolicyoptimizationllm, wang2025ragenunderstandingselfevolutionllm}.

\section{Experiments}
\label{sec:Experiments}
We first evaluate \framework{} on two ``synthetic'' environments (Sokoban and ALFWorld) that require efficient planning for successful task completion.
% These lightweight environments allow us to provide detailed analysis and ablations on the different reasoning styles (e.g., DeepSeek-V3, R1, or \DFS{}) as well as different RL algorithms.
These lightweight environments allow us to provide detailed analysis of the different reasoning styles as well as different RL algorithms.
Then, we extend our methods to a more complex and realistic environment (AndroidWorld).

\subsection{Text Games}
\label{subsec:Toy Text}
% \subsubsection{Experiment Setup}
% \label{subsec:Toy Text}

\paragraph{Benchmarks} Sokoban \citep{SchraderSokoban2018} is a grid-world game where the agent needs to push boxes to target destinations while avoiding obstacles, and successful task completion requires spatial planning to avoid deadlock situations.
ALFWorld \citep{shridhar2021alfworldaligningtextembodied} is a text-based embodied environment where the agent needs to locate/interact with objects to complete household tasks using natural language instructions.
To evaluate the agent's generalization ability, we construct training set, an in-distribution (ID) test set, and an out-of-distribution (OOD) test set.
For Sokoban, we use training set with 6x6 room layouts with 1 box and 1 destination; ID test set with different 6x6 room layouts than training; and OOD test set with 8x8 room layouts with 1 box and 1 destination.
For ALFWorld, we directly use the official training, ID, and OOD test splits from \citet{shridhar2021alfworldaligningtextembodied}.

\paragraph{Baselines setup} To evaluate \DFS{}, we compare against (1) ReACT based prompting methods with models such as GPT-4o \citep{gpt-4o}, Claude-3.7 \citep{claude-3.7}, DeepSeek-V3 \citep{deepseekai2025deepseekv3technicalreport}, and DeepSeek-R1 \citep{deepseekai2025deepseekr1incentivizingreasoningcapability}; and (2) training methods that distill the reasoning traces from strong policy models such as DeepSeek-R1.
To evaluate stage 2 \RLWMshort{} training, we compare against other popular group-based RL algorithms such as RLOO \citep{kool2019buy} and GRPO \citep{shao2024deepseekmathpushinglimitsmathematical}.
Overall, we also compare against \dynathink{} \citep{yu2025dynathinksynergizingreasoningacting}, which similarly uses two-stage training (DIT and DDT) to improve model's simulation ability.

% \paragraph{Implementation Details}
\paragraph{\framework{} setup}
To instantiate \DFS{}, we use Qwen2.5-32B-Instruct \citep{qwen2025qwen25technicalreport} as rollout and value function models, finetuned on rollouts obtained by using DeepSeek-V3 (see \Cref{subsec:DFS Implementation Details} for more details) and use DeepSeek-V3 as the LLM to aggregate the search tree into a single response.
For Sokoban, we use $d=5,b=16,b_{\textrm{train}}=2$; for ALFWorld, we use $d=2,b=24,b_{\textrm{train}}=4$.
We note that all models used by \DFS{} are by themselves much weaker than other models such as DeepSeek-R1 as well as \DFS{} itself.
Since DeepSeek-R1 and \DFS{} have a higher success rate than DeepSeek-V3, to isolate improvement from better reasoning from simply training with more (diverse) data, we thus \emph{only used trajectories where all methods correctly solved the task} for stage 1 training. This results in a total of 207 trajectories in Sokoban and 200 trajectories in ALFWorld from each method (DeepSeek-R1, DeepSeek-V3, and \DFS{}) in the subsequent stage 1 training.

% To instantiate \RLWMshort{}, we continue training the best model from stage 1 distillation. To ensure a fair comparison, we use identical hyperparameters for all methods (RLOO, GRPO, and \RLWMshort{}), when applicable. We use a batch size of 8 tasks per batch, group size of $G=8$, learning rate of 1e-6, and 300 training steps in total for both Sokoban and ALFWorld.
% For \RLWMshort{}, we use $n_\mathcal{T} = 10$ and $n_\pi = 10$ for Sokoban and $n_\mathcal{T} = 10$ and $n_\pi = 20$ for ALFWorld.
% All training are performed on top of Qwen2.5-7B \citep{qwen2025qwen25technicalreport} using 8xH100.
To instantiate \RLWMshort{}, we continue training the best model from stage 1 distillation. To ensure a fair comparison, we use identical hyperparameters for all methods (RLOO, GRPO, and \RLWMshort{}), when applicable. For \RLWMshort{}, we use $n_\mathcal{T} = 10$ and $n_\pi = 10$ for Sokoban and $n_\mathcal{T} = 10$ and $n_\pi = 20$ for ALFWorld.
For more setup details, please see \Cref{subsec:Additional Training Details Text Game}.

\subsubsection{Main Results}
\label{subsubsec:Toy Text Main Results}

\begin{table}[t!]
  \centering
  \caption{Performance on text game environments such as Sokoban and ALFWorld. ``Gen. Token'' denotes the average number of tokens generated per turn. All training in stage-1 and stage-2 are based on Qwen2.5-7B-Instruct. All results are averaged over 3 runs. Our methods are highlighted in \textcolor{gray}{gray}.
  }
  \scalebox{0.82}{
  \begin{tabular}{l l l l ccc ccc}
    \toprule
    \multirow{2}{*}{Method} & 
    % \multirow{2}{*}{Base Model} & 
    \multirow{2}{*}{Gen. Token} & 
    \multicolumn{3}{c}{Sokoban} &
    \multicolumn{3}{c}{ALFWorld}
    \\
    \cmidrule(lr){3-5}
    \cmidrule(lr){6-8}
    & &
    \multicolumn{1}{c}{ID} &
    \multicolumn{1}{c}{OOD} &
    \multicolumn{1}{c}{AVG} &
    \multicolumn{1}{c}{ID} &
    \multicolumn{1}{c}{OOD} &
    \multicolumn{1}{c}{AVG} \\

    \midrule
    %\multicolumn{8}{c}{\emph{Data Collection/Inference}} 
    %\\
    % \cdashline{1-8}
    % \rowcolor{light-gray}
    \react{}(Qwen2.5-7B-Instruct)
    & 1.0x
    & 25.8\tiny{$\pm$1.8} & - & -
    & 35.4\tiny{$\pm$1.9} & - & -\\

    % \rowcolor{light-gray}
    \react{}(Qwen2.5-32B-Instruct)
    & 2.7x
    & 36.7\tiny{$\pm$4.2} & - & -
    & 36.2\tiny{$\pm$3.3} & - & -\\

    % \rowcolor{light-gray}
    \react{}(GPT-4o)
    & 1.5x
    & 37.8\tiny{$\pm$1.0} & - & -
    & 51.3\tiny{$\pm$2.1} & - & -\\

    \react{}(Claude-3.7-Sonnet)
    & 2.3x
    & 70.3\tiny{$\pm$1.2} & - & -
    & 46.1\tiny{$\pm$1.0} & - & -\\

    % \rowcolor{light-gray}
    \react{}(DeepSeek-V3)
    & 2.5x
    & 57.0\tiny{$\pm$1.6} & - & -
    & 55.2\tiny{$\pm$1.0} & - & -\\

    % % \rowcolor{light-gray}
    % - (o4-mini)
    % & 1.0x
    % & 0.0 & - & -
    % & 0.0 & - & -\\

    % \rowcolor{light-gray}
    \react{}(DeepSeek-R1)
    & 14.5x
    & \textbf{96.6}\tiny{$\pm$0.2} & - & -
    & 62.5\tiny{$\pm$0.5} & - & -\\

    \rowcolor{light-gray}
    \DFS{}
    & 2.0x
    & 96.4\tiny{$\pm$0.2} & - & -
    & \textbf{87.7}\tiny{$\pm$1.1} & - & -\\

    % \midrule
    \cmidrule(lr){1-8}
    %\multicolumn{8}{c}{\emph{\dynathink{}}} \\
    \emph{\dynathink{} } \\
    % \cdashline{1-8}
    %\cmidrule(lr){1-1}
    DIT(R1)+DDT($\hat{\mathcal{T}}$)
    & 24.2x
    & 74.0\tiny{$\pm$1.4} & 57.5\tiny{$\pm$1.2} & 65.8\tiny{$\pm$1.9}
    & 63.2\tiny{$\pm$1.5} & 56.7\tiny{$\pm$2.8} & 58.9\tiny{$\pm$2.3}\\

    % \midrule
    %\cmidrule(lr){1-8}
    %\multicolumn{8}{c}{\emph{\framework{}}} \\
    %\cdashline{1-8}
    \cmidrule(lr){1-8}
    \emph{\framework{} Stage 1 (SFT)} \\
    \Distill{}(V3)
    & 2.1x
    & 49.2\tiny{$\pm$1.1} & 34.4\tiny{$\pm$1.3} & 41.8\tiny{$\pm$1.1}
    & 58.9\tiny{$\pm$1.1} & 56.7\tiny{$\pm$1.0} & 57.8\tiny{$\pm$1.2}\\
    \Distill{}(R1)
    & 24.0x
    & \textbf{72.5}\tiny{$\pm$2.9} & \textbf{57.0}\tiny{$\pm$1.9} & \textbf{64.8\tiny{$\pm$2.5}}
    & 59.4\tiny{$\pm$1.5} & 54.2\tiny{$\pm$3.9} & 56.8\tiny{$\pm$3.5}\\
    \rowcolor{light-gray}
    \Distill{}(\DFS{})
    & 2.0x
    & 71.9\tiny{$\pm$1.5} & 55.5\tiny{$\pm$1.6} & 63.7\tiny{$\pm$1.9}
    & \textbf{78.9}\tiny{$\pm$2.1} & \textbf{69.3}\tiny{$\pm$1.3} & \textbf{74.1}\tiny{$\pm$1.8}\\

    % \midrule
    % \cmidrule(lr){2-8}
    \emph{\framework{} Stage 2 (RL)} \\

    \Distill{}(\DFS{}) + RLOO
    & 2.2x
    & 78.1\tiny{$\pm$1.8} & 65.1\tiny{$\pm$1.3} & 71.3\tiny{$\pm$0.9}
    & 85.9\tiny{$\pm$1.3} & 85.4\tiny{$\pm$2.0} & 85.5\tiny{$\pm$2.0}\\
    \Distill{}(\DFS{}) + GRPO
    & 2.1x
    & 79.1\tiny{$\pm$1.3} & 67.8\tiny{$\pm$0.6} & 73.1\tiny{$\pm$1.4}
    & 87.0\tiny{$\pm$3.2} & 87.1\tiny{$\pm$1.1} & 87.0\tiny{$\pm$1.8}\\
    \rowcolor{light-gray}
    \Distill{}(\DFS{}) + \RLWMshort{}
    & 1.9x
    & \textbf{82.5}\tiny{$\pm$1.5} & \textbf{70.1}\tiny{$\pm$1.6} & \textbf{77.1}\tiny{$\pm$1.7}
    & \textbf{92.5}\tiny{$\pm$0.8} & \textbf{89.1}\tiny{$\pm$1.3} & \textbf{90.8}\tiny{$\pm$0.9}\\
    \bottomrule
\end{tabular}
  }
  \vspace{-3mm}
  \label{tbl:main_rl_result}
\end{table}
% \input{floats/main_rl_result}

% \paragraph{\DFS{} Results}
In the upper section of \Cref{tbl:main_rl_result}, we first evaluate \DFS{}'s performance against other strong reasoning models such as DeepSeek-R1.
Then, we compare different training methods to integrate/improve the simulation ability of the policy model.
In \Cref{tbl:main_rl_result}, we first find that \DFS{} achieves near-perfect performance on Sokoban (96.4\% success) and a strong performance on ALFWorld (87.7\% success), significantly outperforming all other methods.
On Sokoban, we find strong reasoning models such as DeepSeek-R1 also achieves near-perfect performance, which we attribute to R1's ability to correctly simulate Sokoban game states (but not on ALFWorld) during its reasoning process (see \Cref{subsubsec:World Model Accuracy} for empirical results).
% Our method, by construction, uses ground-truth simulations during its reasoning process, hence achieving strong performance in both environments.
In contrast, \DFS{} utilizes ground-truth simulations from search trees, and hence was able to achieve strong performance in both environments.

% In stage 1 training, we find that strong performance from \DFS{} can be learned by SFT training: \Distill{}(\DFS{}) achieves a similar performance to \Distill{}(R1) on Sokoban but significantly outperforms both \Distill{}(V3) and \Distill{}(R1) on ALFWorld.
% Additionally, since \DFS{} constructs reasoning traces consists almost entirely of only planning via simulation (see \Cref{fig:dfs_algo_fig} Left), \Distill{}(\DFS{}) outputs \emph{11x less tokens} on average compared to \Distill{}(R1).
% These results indicate that reasoning traces that contain high-quality simulation traces are crucial for performance in planning-intensive, long-horizon tasks.
In stage 1 training, we find \Distill{}(\DFS{}) achieves a similar performance to \Distill{}(R1) on Sokoban but significantly outperforms both \Distill{}(V3) and \Distill{}(R1) on ALFWorld.
Additionally, since \DFS{} constructs reasoning traces consists almost entirely of only planning via simulation (see \Cref{fig:dfs_algo_fig} Left), \Distill{}(\DFS{}) outputs \emph{11x less tokens} on average compared to \Distill{}(R1).
These results indicate that that strong performance from \DFS{} can be learned by SFT, and that the ability to model and simulate the environment is crucial for long-horizon, planning-intensive tasks.

% \paragraph{\RLWMshort{} Results}
In stage 2 training, we continue from the best model (\Distill{}(\DFS{})) with online RL. In \Cref{tbl:main_rl_result}, we find that \RLWMshort{} improves upon both GRPO, RLOO, as well as \dynathink{}, while maintaining a similar output token length compared to its base model \Distill{}(\DFS{}).
This indicates that \RLWMshort{} is effective at improving the model's simulation ability during online RL training (also see \Cref{subsubsec:World Model Accuracy} for empirical results), and that improving such simulation ability helps improve task performance.

\begin{table}[t!]
    \centering
    \caption{Measuring simulation ability of different models across different training stages. We report the average success rate and the simulation ability (Sim Score $\in [0, 1]$) averaged across all trajectories. We also report the correlation coefficient $r$ between the success rate and the simulation score.}
    \vspace{-2mm}
    \scalebox{0.9}{
    % \begin{tabular}{l l ccc ccc}
    \begin{tabular}{l l cc cc}
      \toprule
      \multirow{2}{*}{Method} & 
      \multicolumn{2}{c}{Sokoban} &
      \multicolumn{2}{c}{ALFWorld}
      \\
      \cmidrule(lr){2-3}
      \cmidrule(lr){4-5}
      & 
      \multicolumn{1}{c}{Success} &
      \multicolumn{1}{c}{Sim Score} &
      \multicolumn{1}{c}{Success} &
      \multicolumn{1}{c}{Sim Score} \\
      \midrule
      % \emph{Data Collection/Inference} \\
      % \multicolumn{5}{c}{\emph{Data Collection/Inference}} \\
      \react{}(Qwen2.5-7B-Instruct)
      & 25.8\tiny{$\pm$1.8} & 0.21\tiny{($r=$0.64)}
      & 35.4\tiny{$\pm$1.9} & 0.18\tiny{($r=$0.46)}\\

      \react{}(DeepSeek-V3)
      & 57.0\tiny{$\pm$1.6} & 0.54\tiny{($r=$0.81)}
      & 55.2\tiny{$\pm$1.0} & 0.35\tiny{($r=$0.68)}\\

      \react{}(DeepSeek-R1)
      & \textbf{96.6}\tiny{$\pm$0.2} & 0.93\tiny{($r=$0.96)}
      & 62.5\tiny{$\pm$0.5} & 0.36\tiny{($r=$0.70)}\\

      \rowcolor{light-gray}
      \DFS{}
      & 96.4\tiny{$\pm$0.2} & \textbf{1.00}\tiny{(-)}\phantom{aaaaaaaa}
      & \textbf{87.7}\tiny{$\pm$1.1} & \textbf{1.00}\tiny{(-)}\phantom{aaaaaaaa}\\

      \midrule
      % \multicolumn{5}{c}{\emph{Dyna-Think}} \\
      \emph{Dyna-Think} \\
      DIT(R1)+DDT($\hat{\mathcal{T}}$)
      & 74.0\tiny{$\pm$1.4} & 0.62\tiny{($r=$0.74)}
      & 63.2\tiny{$\pm$1.5} & 0.36\tiny{($r=$0.76)}\\

      \midrule
      % \multicolumn{5}{c}{\emph{\framework{}}} \\
      \emph{\framework{} Stage 1 (SFT)} \\
      \Distill{}(R1)
      & \textbf{72.5}\tiny{$\pm$2.9} & 0.61\tiny{($r=$0.75)}
      & 59.4\tiny{$\pm$1.5} & 0.34\tiny{($r=$0.77)}\\

      \rowcolor{light-gray}
      \Distill{}(\DFS{})
      & 71.9\tiny{$\pm$1.5} & \textbf{0.62}\tiny{($r=$0.78)}
      & \textbf{78.9}\tiny{$\pm$2.1} & \textbf{0.37}\tiny{($r=$0.74)}\\

      % \midrule
      \emph{\framework{} Stage 2 (RL)} \\
      \Distill{}(\DFS{}) + GRPO
      & 79.1\tiny{$\pm$1.3} & 0.62\tiny{($r=$0.65)}
      & 87.0\tiny{$\pm$3.2} & 0.38\tiny{($\rho=$0.48)}\\

      \rowcolor{light-gray}
      \Distill{}(\DFS{}) + \RLWMshort{}
      & \textbf{82.5}\tiny{$\pm$1.5} & \textbf{0.67}\tiny{($r=$0.64)}
      & \textbf{92.5}\tiny{$\pm$0.8} & \textbf{0.43}\tiny{($r=$0.55)}\\
      \bottomrule
  \end{tabular}
    }
    \vspace{-5mm}
    \label{tbl:wm_acc_result}
\end{table}
\subsubsection{Measuring Simulation Ability}
\label{subsubsec:World Model Accuracy}

% In this work, we introduced \DFS{} and \RLWMshort{} to integrate and improve the simulation ability of agents during reasoning.
\framework{} aims to integrate and improve the simulation ability of agents.
To measure this simulation ability, we evaluate the \textbf{Simulation Score} (Sim Score) of different models and the \textbf{Spearman Correlation Coefficient} ($r_s$) between sim score and success rate.
Given a state $s_t$ and generated response $a_{t} \sim \pi_\theta(\cdot|s_t)$, we evaluate the simulation score of $a_{t}$ by 1) first prompting an LLM to extract the final action plan $(\hat{a}_1, \hat{a}_2, \cdots, \hat{a}_d)$ and the natural language description (i.e., simulation) of the corresponding imagined next-states $(\hat{s}_{t+1}, \hat{s}_{t+2}, \cdots, \hat{s}_{t+d})$ from the response $a_{t}$; 2) then execute the action plan in the environment to obtain ground truth next-states $\{s_{t+1}, s_{t+2}, \cdots, s_{t+d}\}$; 3) finally, prompt an LLM to judge \citep{zheng2023judgingllmasajudgemtbenchchatbot} the correctness of these simulated next-states $\hat{s}_i$ by comparing them against the ground truth $s_i$, returning a score $\in [0, 1]$. Finally, we averaged the score for each turn to obtain an overall simulation score for the trajectory. To ensure a fair judgment, we used a different LLM from all of our experiments (Qwen3-235B-A22B-Instruct \citep{qwen3technicalreport}). For judgment prompts, please see \Cref{subsec:World Model Score Prompts}.

We present the results in \Cref{tbl:wm_acc_result}. In \Cref{tbl:wm_acc_result}, we find that 1) \DFS{} maintains its strong success rates across both Sokoban and ALFWorld due to its perfect simulation ability (by construction), whereas DeepSeek-R1 struggled in ALFWorld as it struggles to model the environment layout;
and 2) both \Distill{}(\DFS{}) and \RLWMshort{} improve the simulation ability alongside task performance compared to their baselines.
These results show that our methods helped improve the simulation ability of the model beyond simply improving task performance.
% We now evaluate changes of world model performance across different training stages. To evaluate world modeling quality during reasoning, we designed a simple rubric and prompted a strong model not used throughout the data collection/training process to provide a more objective evaluation. For each turn in a trajectory, we measure the world modeling quality using an LLM and then produce a trajectory-level score by averaging scores of all turns. We report the result in \Cref{tbl:wm_acc_result}.

% \vspace{-2mm}
\subsection{AndroidWorld}
\label{subsec:AndroidWorld}
Next, we extend our \framework{} to AndroidWorld \citep{rawles2025androidworlddynamicbenchmarkingenvironment} - a highly challenging benchmark that evaluates the agent's ability control and complete tasks on a virtual Android device.

%\subsubsection{Experiment Setup}
%\label{subsubsec:AndroidWorld Setup}
\vspace{-2.5mm}
\paragraph{Benchmarks} AndroidWorld \citep{rawles2025androidworlddynamicbenchmarkingenvironment} provides a fully functional Android environment that requires the agent to interact with Android's GUI to complete tasks across 20 real-world Android apps.
Since tasks in AndroidWorld are parameterized by task types (116), we construct a training set with 81 task types with in total 1946 tasks, an ID test set with 128 different tasks from the same task types, and an OOD test set with 128 tasks from the remaining 35 held-out task types.
We use a maximum number of 15 steps and the screenshot-only modality as input.
We provide an example task and action in \Cref{subsec:Example Task and Actions in AndroidWorld}.

\vspace{-2.5mm}
\paragraph{Baselines setup}
Since our methods consider end-to-end training, we compare against models that are capable of directly generating executable actions given an GUI screenshot, and exclude modular systems such as \citet{gou2025navigatingdigitalworldhumans,agashe2025agents2compositionalgeneralistspecialist}.
We thus mainly compare against (1) \react{} based prompting method with Qwen2.5-VL-72B/7B \citep{bai2025qwen25vltechnicalreport}, and GPT-4o; and (2) distillation from Qwen2.5-VL-72B\footnote{
    We were unable to reproduce the reported performance of more recent GUI models such as UI-Tars1.5 \citep{qin2025uitarspioneeringautomatedgui}, and hence focus on using Qwen2.5-VL for simplicity. Please see \Cref{subsec:Implementation/Evaluation Details} for more details.
}. 
To evaluate stage 2 \RLWMshort{}, we compare against GRPO following \Cref{subsec:Toy Text}.
We exclude comparison against \dynathink{} in this experiment, because DDT($\hat{\mathcal{T}}$) trains the model to predict next-state (in this case, screenshot images), which cannot be implemented using most VLMs as they can only generate text.

% \paragraph{Implementation Details}
\paragraph{\framework{} setup}
% We note that the model used to synthesize data (DeepSeek-V3 and Qwen2.5-32B distilled from DeepSeek-V3) are by themselves much weaker than the final DFS performance.
Since AndroidWorld is a highly challenging and compute-intensive environment (each episode on average takes 15-20 minutes to complete), we do not perform any rollout/value function training for \DFS{}. Instead, we directly prompt Qwen2.5-VL-72B as the rollout model, prompt GPT-4o as a judge to approximate the value function, and also use GPT-4o as the VLM to aggregate the rollouts into a single response in \DFS{}.
We use $d=1,b=16,b_{\textrm{train}}=4$ for \DFS{}, and a total of 128 trajectories for distillation/stage 1 training. To instantiate \RLWMshort{}, we generally followed the same recipe as \Cref{subsec:Toy Text}, but used less training steps (60) as AndroidWorld is highly compute-intensive and time-consuming.
For more details, please see \Cref{subsec:Additional Training Details AndroidWorld}.

\begin{table}[t!]
    \centering
    \caption{Performance on AndroidWorld. All training in stage-1 and stage-2 are based on Qwen2.5-VL-7B/32B-Instruct.
    We exclude \dynathink{} since (most) VLMs cannot predict \emph{images}, as required by DDT($\hat{\mathcal{T}}$) training.
    All results are averaged over 3 runs. Our methods are highlighted in \textcolor{gray}{gray}.
    }
    % \vspace{-2mm}
    \scalebox{0.9}{
    \begin{tabular}{l l l l ccc}
      \toprule
      \multirow{2}{*}{Method} & 
      % \multirow{2}{*}{Base Model} & 
      \multirow{2}{*}{Gen. Token} & 
      \multicolumn{3}{c}{AndroidWorld} \\
      \cmidrule(lr){3-5}
      & &
      \multicolumn{1}{c}{ID} &
      \multicolumn{1}{c}{OOD} &
      \multicolumn{1}{c}{AVG} \\
      \midrule
      % \emph{Data Collection/Inference} \\
      % \multicolumn{5}{c}{\emph{Data Collection/Inference}} \\
      \react{}(GPT-4o)
      & 1.0x
      & 5.1\tiny{$\pm$0.2} & - & -\\

      % \rowcolor{light-gray}
      \react{}(Qwen2.5-VL-7B-Instruct)
      & 1.0x
      & 5.3\tiny{$\pm$0.2} & - & -\\
  
      % \rowcolor{light-gray}
      \react{}(Qwen2.5-VL-72B-Instruct)
      & 1.1x
      & 19.5\tiny{$\pm$0.4} & - & -\\
  
      \rowcolor{light-gray}
      \DFS{}
      & 2.1x
      & \textbf{34.4}\tiny{$\pm$0.4} & - & -\\
  
      \midrule
      % \multicolumn{5}{c}{\emph{\framework{}}} \\
      \emph{\framework{} Stage 1 (SFT)} \\
      \Distill{}-7B(Qwen2.5-VL-72B-Instruct)
      & 1.0x
      & 13.1\tiny{$\pm$0.4} & 8.6\tiny{$\pm$0.2} & 10.8\tiny{$\pm$0.6}\\
      \rowcolor{light-gray}
      \Distill{}-7B(\DFS{})
      & 2.1x
      & {21.1}\tiny{$\pm$0.4} & {10.2}\tiny{$\pm$0.6} & {15.7}\tiny{$\pm$0.8}\\
      \rowcolor{light-gray}
      \Distill{}-32B(\DFS{})
      & 2.0x
      & \textbf{32.8}\tiny{$\pm$0.4} & \textbf{15.6}\tiny{$\pm$0.7} & \textbf{24.2}\tiny{$\pm$0.6}\\
  
      % \midrule
      % \cdashline{1-5}
      \emph{\framework{} Stage 2 (RL)} \\
      % \rowcolor{light-gray}
      % \Distill{}(\DFS{}) + \RLWMshort{}
      % & 2.2x
      % & \textbf{25.2}\tiny{$\pm$0.6} & \textbf{15.6}\tiny{$\pm$0.8} & \textbf{20.4}\tiny{$\pm$1.3}\\
      \Distill{}-32B(\DFS{}) + GRPO
      & 2.1x
      & 35.3\tiny{$\pm$0.4} & 20.3\tiny{$\pm$0.6} & 27.8\tiny{$\pm$0.4}\\
      \rowcolor{light-gray}
      \Distill{}-32B(\DFS{}) + \RLWMshort{}
      & 1.9x
      & \textbf{40.7}\tiny{$\pm$1.0} & \textbf{22.9}\tiny{$\pm$1.0} & \textbf{31.8}\tiny{$\pm$1.0}\\
      \bottomrule
  \end{tabular}
    }
    % \vspace{-2mm}
    \label{tbl:android_rl_result}
\end{table}

% \vspace{-2mm}
\subsubsection{Main Results}
\label{subsubsec:AndroidWorld Main Results}
\paragraph{Results} We present the results in \Cref{tbl:android_rl_result}. In general, we observe similar results compared to \Cref{subsubsec:Toy Text Main Results}. First, we find that \DFS{} inference significantly improves performance, and that the improved performance can be transferred to Qwen2.5-VL-7B and 32B via \Distill{}(\DFS{}).
Next, in both training stages of \framework{}, we find improved performance in both ID and OOD test sets compared to baselines, including Qwen2.5-VL-72B and even \DFS{}.
These results highlight the effectiveness of our method to improve agent's performance in complex environments.

%\subsubsection{Error Analysis}
% \vspace{-2mm}
\paragraph{Error Analysis}
\label{subsubsec:AndroidWorld Error Analysis}
Compared to synthetic text games (\Cref{subsubsec:Toy Text Main Results}) where \DFS{} achieves near-perfect performance, we find \DFS{} struggles in AndroidWorld despite improvements compared to baselines.
% After analyzing trajectories produced by \DFS{}, we find performance is bottlenecked the by the weak ability of the rollout model (i.e., Qwen2.5-VL-72B), primarily due to 1) incorrect/incomplete understanding of the current interface and the functionality of certain buttons; and 2) inability to significantly change its course of action after making more than one mistakes.
After analyzing trajectories produced by \DFS{}, we find performance is bottlenecked by the rollout model (Qwen2.5-VL-72B), mainly due to: (1) incomplete understanding of some GUI interfaces and certain button functions, and (2) inability to recover after making multiple mistakes.
We believe methods to improve the foundation model's capability could mitigate these problems \citep{wang2025opencuaopenfoundationscomputeruse,qin2025uitarspioneeringautomatedgui}, which we leave for future work.

% \section{Analysis}
% \label{sec:Analysis}

% \subsection{Test-time scaling}
% \label{subsec:Test-time scaling}
% very easy to run, and has some preliminary results

% \vspace{-2mm}
\section{Conclusion}
\label{sec:Conclusion}
% \vspace{-1mm}
In this work, we propose \framework{} to synergize reasoning with simulations for autonomous AI agents.
We empirically show that an agent's ability to model and simulate the environment strongly correlates with its ability to correctly reason and complete long-horizon, planing-intensive tasks.
We introduce \framework{}, a two-stage training method to explicitly teach (V)LM
agents to integrate and improve such simulation a part of their reasoning.
% In stage 1 training, we propose \DFS{} to algorithmically construct reasoning data using ground-truth future outcomes (i.e., states and actions) obtained by real environment interactions, and subsequently trained the agent to imitate these reasoning traces.
In stage 1 training, we propose \DFS{} to train a model to simulate
future states by learning to predict an expanded search tree in their reasoning.
In stage 2 training, we propose \RLWMshort{} to further refine the agent's reasoning and simulation ability using online RL.
Empirical results on three benchmarks show that 
(1) \DFS{} effectively teaches AI agents to simulate; and (2) \RLWMshort{}, by leveraging both outcome rewards and intermediate interactions, learns better policies for long-horizon, planning-intensive tasks.
% 1) reasoning traces constructed by \DFS{} is highly effective at improving agent's performance on long-horizon, planning-intensive tasks; and 2) world modeling improvements achieved by \framework{} translates to the model's improved ability to solve complex agentic tasks.

% \subsubsection*{Acknowledgments}
% Use unnumbered third level headings for the acknowledgments. All

\bibliography{iclr2026_conference}
\bibliographystyle{iclr2026_conference}

\clearpage
\appendix

\setcounter{table}{0}
\renewcommand{\thetable}{A\arabic{table}}
\setcounter{figure}{0}
\renewcommand{\thefigure}{A\arabic{figure}}

\section{LLM Usage}
% general purpose writing assistant tool.
% This work used LLMs as general purpose writing assistant tools, mainly to improve the grammar and wording of this paper to improve clarity.
This work used LLMs as general-purpose writing assistants to improve the grammar and clarity of the paper.
We {\it did not} use LLMs to generate any research ideas, automate experiments, or analyze results.

\section{Ethics Statement}
\label{label:ethics statement}
Generally, while most methods and models are not designed for unethical usage, there is often potential for abuse in their applications.
Autonomous AI agents can be used for a variety of tasks such as automating information gathering, software development, computer/phone-use and more.
In this work, we proposed our \framework{} framework to enhance the simulation ability and hence performance of AI agents.
However, since AI agents are fundamentally task-agnostic, it is possible to use them for unethical tasks such as scamming or disseminating false information on the internet.
We believe developing guardrails such as safety filters \citep{openai-content-filters,inan2023llamaguardllmbasedinputoutput} are highly valuable for AI agent research.
We do not condone the \framework{} or its constituent methods for any unlawful or morally unjust purposes.

\section{Additional Algorithmic Details}
\label{sec:Additional Algorithmic Details}

In \Cref{algo:wmimp}, we provide the pseudo-code for \wmrolloutshort{}. On a high level, \wmrolloutshort{} aims to generate a refined response at a given state $s_t$ with better simulation content compared to that of the original response. Specifically, \wmrolloutshort{} first performs normal inference $a_{t} \sim \pi_\theta(\cdot|s_t)$ to generate a response; extracts the plan $(\hat{a}_1, \hat{a}_2, \cdots, \hat{a}_d)$ from $a_{t}$ using the ``<plan></plan>'' tags (see \Cref{tab:sokoban_example} for example response with such tags); executes the extracted plan in the environment and obtain the actual next-states $\{s_{t+1}, s_{t+2}, \cdots, s_{t+d}\}$; and finally, prompts an LLM to refine the original response based on the actual next-states, using the prompt in \Cref{tab:wmrefine_prompt}. The resulting refined response $a^{\textrm{refine}}_{t}$ is then used as the next action $a_{t}$, and this process is repeated until the task is completed or a maximum number of steps is reached.

\begin{algorithm}[H]
    \caption{Simulation Refinement Rollout (\wmrolloutshort{})}\label{algo:wmimp}
    \begin{algorithmic}[1]
    \Require policy $\pi_\theta$, environment $\mathcal{T}$, group size $G$
    \State repeat the following $G$ times:
    \State $\tau' \gets \{\}, \tau'_{\textrm{refine}} \gets \{\}, t=0, s_{0} \gets T$
    \While {not done and $t <t_{\max}$}
        \State $a \gets \pi_\theta(s_{t})$
        \State $\{\hat{a}_{1}, \cdots, \hat{a}_{n}\} \gets$ extract\_plan($a$)
        \State \emph{\textcolor{gray}{// improve action $a$ using next-state information}}
        \State $\{s_{t+1}, \cdots, s_{t+n}\} \gets \{\mathcal{T}(s_{t}, \hat{a}_{1}), \cdots, \mathcal{T}(s_{t+n-1}, \hat{a}_{n})\}$
        % \State $s^{imp}_t \gets$ selfimp\_prompt($s_t, a, \{s_{t+1}, a'_{1}, \cdots, s_{t+n}\}$)
        \State $s^{\textrm{refine}}_t \gets$ refinement prompt($a | s_t, a, \{s_{t+1}, \hat{a}_{1}, \cdots, s_{t+n}\}$)  \emph{\textcolor{gray}{// see \Cref{tab:wmrefine_prompt}}}
        \State $a^{\textrm{refine}} \gets \pi_\theta(s^{\textrm{refine}}_t)$
        \State \emph{\textcolor{gray}{// update episode buffer}}
        \State $\tau' \gets \tau' \cup \{s_t, a^{\textrm{refine}}\}$  \emph{\textcolor{gray}{// learn improved policy}}
        \State $\tau'_{\textrm{refine}} \gets \tau'_{\textrm{refine}} \cup \{s^{\textrm{refine}}_{t}, a^{\textrm{refine}}\}$ \emph{\textcolor{gray}{// learn to refine simulations}}
        \State $s_{t+1} \gets \mathcal{T}(s_{t}, a^{\textrm{refine}})$
        \State $t \gets t + 1$
    \EndWhile
    \State \Return $\tau', \tau'_{\textrm{refine}}$
    \end{algorithmic}
\end{algorithm}

\section{Additional Details on Text Games}
\label{sec:Additional Details on Text Games}

\subsection{Example Tasks and Actions}
\label{subsec:Example Tasks and Actions in Text Games}
Sokoban \citep{SchraderSokoban2018} is a grid-world game where the agent needs to push boxes to their destinations while avoiding obstacles.
Valid actions in Sokoban are up, down, left, and right.
As an example, we provide an example input state and generated action in \Cref{tab:sokoban_example}.
ALFWorld \citep{shridhar2021alfworldaligningtextembodied} is a text-based embodied environment where the agent needs to locate/interact with objects to complete embodied household tasks using natural language instructions.
Valid actions in ALFWorld are dependent on what's available in the current state.
We provide an example input state and generated action in \Cref{tab:alfworld_example}.

\subsection{\DFS{} Implementation Details}
\label{subsec:DFS Implementation Details}

We provide a pseudo-code for \DFS{} in \Cref{algo:ourdfs}.
% How rollout and value functions are trained
For text games, we finetune Qwen2.5-32B-Instruct as rollout and value function models using DeepSeek-V3's rollouts.
Specifically, we first use DeepSeek-V3 to generate 256 rollouts using tasks from the training set.
Then, to train the rollout model, we simply perform SFT training on one correct rollout for each task.
To train the value function, we use the trained policy model to generate the same 256 rollouts, repeated over 3 times, and compute $V(s_t)$ as the probability of successfully completing the task from $s_t$ across all trajectories that contains $s_t$, discounted by the number of remaining steps needed in the current trajectory:
\[
    V(s_t) = \gamma^{t_{\max}-t} \frac{1}{|\mathrm{T}|} \sum_{\tau \in \mathrm{T}} \mathbbm{1}[\tau \text{ is successful}], \quad \text{where }\mathrm{T} \equiv \{ \tau_1, \tau_2, \cdots | s_{t} \in \tau_i\}
\] 
where $\gamma$ is the discount factor and $t_{\max}$ is the maximum number of steps in a trajectory.
In both environments, we used $\gamma = 0.95$.
Finally, we finetune a separate Qwen2.5-32B-Instruct as the value function by adding a linear value head to the model architecture, and perform MSE loss training on the computed $V(s_t)$ across all states from all trajectories.

Since Sokoban and ALFWorld environments are fast, these rollouts were completed within 1 hour. For complex environments such as AndroidWorld, we directly prompt pretrained VLMs such as Qwen2.5-VL-72B and GPT-4o as rollout and value function models (\Cref{subsec:AndroidWorld}).

\subsection{Simulation Refinement Performance}
\label{subsec:World Model Refinement Performance}

To empirically show that (V)LMs are capable of leveraging next-state information to improve their action, we evaluate the performance of \wmrolloutshort{} compared to direct prompting (\react{}).
We report the result in \Cref{tbl:wmimp_algo_result}.

In general, we find that 1) all models showed improved task success rate when provided with next-state information; and 2) stronger models such as GPT-4o and GPT-4.1 \citep{gpt-4o, gpt-41} shows larger improvement compared to weaker models such as Qwen2.5-7B-Instruct.
We believe this is because correcting its own mistakes is requires non-trivial reasoning ability, which is more difficult for weaker models such as Qwen2.5-7B-Instruct to achieve.
Overall, this result indicates that world modeling error (e.g., especially for tasks such as ALFWorld) remains a significant bottleneck for (V)LM agents reasoning ability in long-horizon tasks.

\begin{table}[t!]
  \centering
  \caption{\wmrolloutshort{} performance on Sokoban and ALFWorld. We show that when provided with ground-truth next-state information (\wmrolloutshort{}), models \emph{achieve better performance} compared to direct prompting (\react{}).
  }
  \scalebox{0.9}{
  \begin{tabular}{l l l l ccc ccc}
    \toprule
    \multirow{1}{*}{Base Model} & 
    \multirow{1}{*}{Method} & 
    \multicolumn{1}{c}{Sokoban} &
    \multicolumn{1}{c}{ALFWorld}
    \\
    \midrule
    % \rowcolor{light-gray}
    Qwen2.5-7B-Instruct & \react{}
    & 25.8\tiny{$\pm$1.8}
    & 35.4\tiny{$\pm$1.9} \\
    
    \rowcolor{light-gray}
    % \wmrolloutshort{} & Qwen2.5-7B-Instruct
     & \wmrolloutshort{}
    & \textbf{30.0}\tiny{$\pm$1.4}
    & \textbf{39.1}\tiny{$\pm$1.6} \\

    % \rowcolor{light-gray}
    % - (ReACT) & GPT-4o-2024-11-20
    GPT-4o-2024-11-20 & \react{}
    & 37.8\tiny{$\pm$1.0}
    & 51.3\tiny{$\pm$2.1} \\

    \rowcolor{light-gray}
    % \wmrolloutshort{} & GPT-4o-2024-11-20
      & \wmrolloutshort{}
    & \textbf{41.4}\tiny{$\pm$1.2}
    & \textbf{64.8}\tiny{$\pm$2.5} \\

    GPT-4.1 & \react{}
    & 67.9\tiny{$\pm$1.0}
    & 54.4\tiny{$\pm$2.1} \\

    \rowcolor{light-gray}
      & \wmrolloutshort{}
    & \textbf{71.1}\tiny{$\pm$1.3}
    & \textbf{67.9}\tiny{$\pm$2.0} \\
    \bottomrule
\end{tabular}
  }
  % \vspace{-5pt}
  \label{tbl:wmimp_algo_result}
\end{table}
\begin{algorithm}[t]
    \caption{\DFS{}}\label{algo:ourdfs}
    \begin{algorithmic}[1]
    \Require policy $\pi_\theta$, value function $V_\nu$, environment $\mathcal{T}$, (V)LM $M$
    \Require hyperparameters $b, d, t_{\max}, b_{\textrm{train}}$
    \State $\tau \gets \{\}, t=0, s_{0} \gets T$
    \While {not done and $t <t_{\max}$}
        \State $\{ \tau^{i} \}_{i=1}^{b} \gets$ sample $b$ rollouts using $\pi_\theta$ starting from $s_t$ for max $d$ steps
        \State $\{ \tau^{i} \}_{i=1}^{b'} \gets$ deduplicate $\{ \tau^{i} \}_{i=1}^{b}$
        \State $\{v^{i}\}_{i=1}^{b'} \gets$ estimate value $\{V_\nu(s^{i}_{t+d})\}_{i=1}^{b}$
        \State \emph{\textcolor{gray}{// subsample rollouts}}
        \State $\tau^{*} \gets \tau^{\argmax_{i} v^{i}}$
        \State $\{\tau^{i}\}_{i=1}^{b_{\textrm{train}}} \gets$ $\{\tau^{*}\} \cup$ subsample $b_{\textrm{train}}-1$ rollouts from the rest of $\{\tau^{i}\}_{i=1}^{b'}$
        \State \emph{\textcolor{gray}{// aggregate rollouts into a single reasoning response}}
        \State $\{\text{plan}^{i}\}_{i=1}^{b_{\textrm{train}}} \gets$ summarize $\{M(\tau^{i}, v^{i})\}_{i=1}^{b_{\textrm{train}}}$
        \State $a^{\DFS{}} \gets $ aggregate $M(s_t, \{\text{plan}^{i}\}_{i=1}^{b_{\textrm{train}}})$
        \State \emph{\textcolor{gray}{// next step}}
        \State $s_{t+1} \gets \mathcal{T}(s_{t}, a^{\DFS{}})$
        \State $\tau \gets \tau \cup \{s_t, a^{\DFS{}}\}$
        \State $t \gets t + 1$
    \EndWhile
    \State \Return $\tau$
    \end{algorithmic}
\end{algorithm}

\subsection{Additional Training Details}
\label{subsec:Additional Training Details Text Game}

To instantiate \RLWMshort{}, we continue training the best model from stage 1 distillation. To ensure a fair comparison, we use identical hyperparameters for all methods (RLOO, GRPO, and \RLWMshort{}), when applicable. We use a batch size of 8 tasks per batch, group size of $G=8$, learning rate of 1e-6, and 300 training steps in total for both Sokoban and ALFWorld.
For \RLWMshort{}, we use $n_\mathcal{T} = 10$ and $n_\pi = 10$ for Sokoban and $n_\mathcal{T} = 10$ and $n_\pi = 20$ for ALFWorld.
All training are performed on top of Qwen2.5-7B \citep{qwen2025qwen25technicalreport} using 8xH100.

\subsection{Simulation Score Prompts}
\label{subsec:World Model Score Prompts}
To evaluate the simulation ability of a model $\pi_\theta$, we use LLM-as-a-judge \citep{zheng2023judgingllmasajudgemtbenchchatbot} to measure the correctness and quality of the simulation generated by $\pi_\theta$ at each turn in a given trajectory.
Specifically, for each $a_{t} \sim \pi_\theta(\cdot|s_t)$, we first prompt an LLM to extract the final action plan $(\hat{a}_1, \hat{a}_2, \cdots, \hat{a}_d)$ from $a_{t}$ and the corresponding natural language description of the next-states $(\hat{s}_{t+1}, \hat{s}_{t+2}, \cdots, \hat{s}_{t+d})$ from the response $a_{t}$. We present the prompts used for Sokoban and ALFWorld in \Cref{tab:sokoban_extract_prompt,tab:alfworld_extract_prompt}, respectively.
Then, we execute the action plan in the environment to obtain ground truth next-states $\{s_{t+1}, s_{t+2}, \cdots, s_{t+d}\}$.
Finally, we prompt an LLM to judge the quality of the plan by comparing ``imagined'' next-states generated by $\pi_\theta$ against the ground truth next-states, using prompts in \Cref{tab:sokoban_judge_prompt,tab:alfworld_judge_prompt}.
This results in a score $\in [0, 1]$ for each turn in the trajectory, which is then averaged across all turns to obtain an overall simulation score for the entire trajectory.

\begin{figure}[t!]
    \centering
    \includegraphics[scale=0.53]{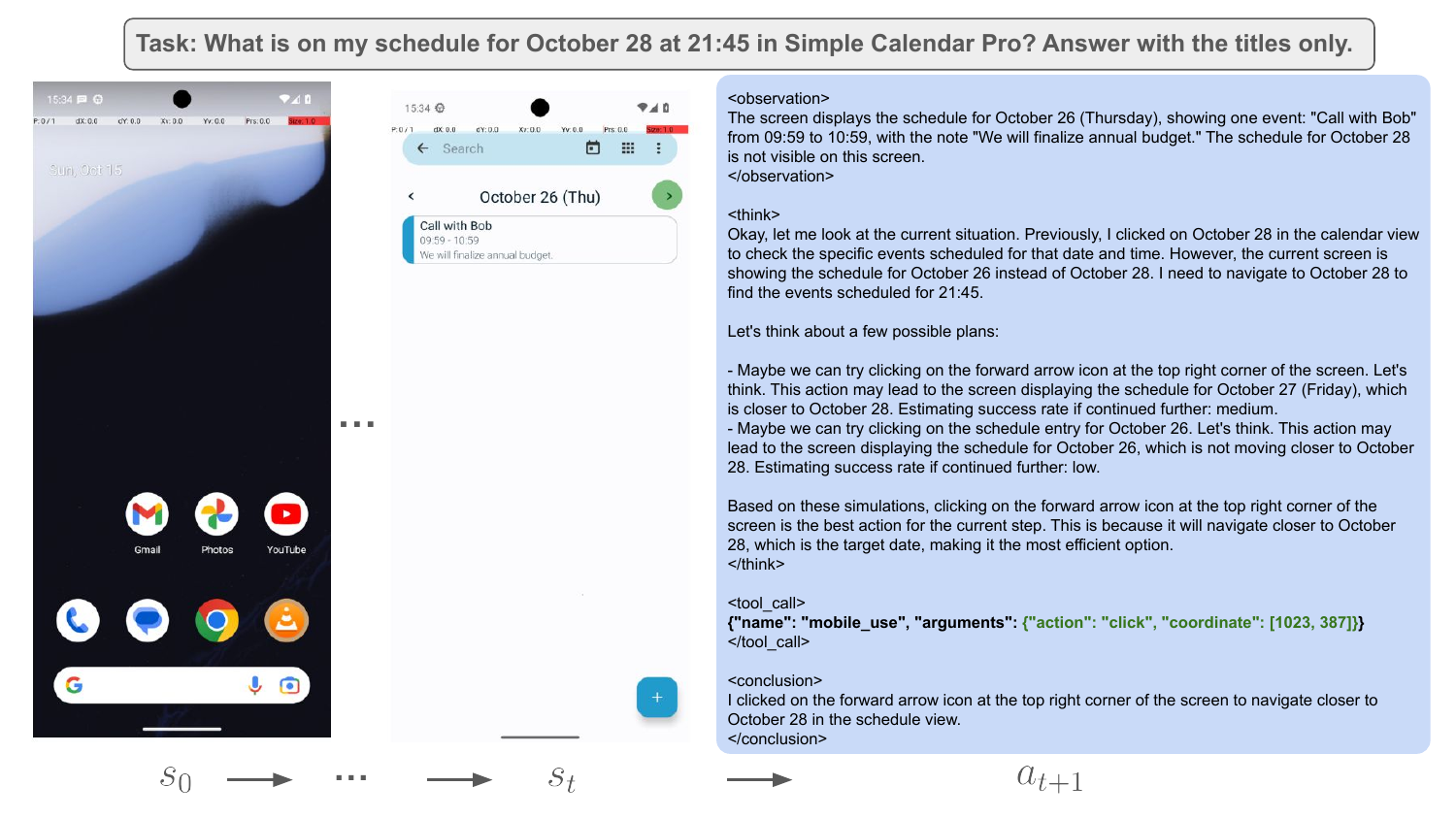}
    \caption{Example task, input screenshot, and output generated by model trained using \framework{}. For clarity, we directly rendered the proposed action in $a_{t+1}$ (click at 1023,387) in green on $s_t$.}
    \label{fig:android_example}
\end{figure}

\section{Additional Details on AndroidWorld}
\label{sec:Additional Details on AndroidWorld}

\subsection{Example Task and Actions in AndroidWorld}
\label{subsec:Example Task and Actions in AndroidWorld}

In this work, we use the dockerized environment provided by AndroidWorld to evaluate and train all methods.
We use the screenshot-only modality.
In \Cref{fig:android_example}, we present an example task, input screenshot $s_t$ from AndroidWorld, as well as an example output $a_{t}$ generated by models trained using \framework{}.
For more details on AndroidWorld, please refer to \citet{rawles2025androidworlddynamicbenchmarkingenvironment}.

\subsection{Additional Training Details}
\label{subsec:Additional Training Details AndroidWorld}

To standardize training and evaluation, we use the dockerized version of AndroidWorld and adapt the action space provided by \citet{rawles2025androidworlddynamicbenchmarkingenvironment}.

To instantiate \RLWMshort{}, we continue training the best model from stage 1 distillation. We followed \Cref{subsec:Toy Text} and used a batch size of 8 tasks per batch, group size of $G=8$, learning rate of 1e-6.
Since AndroidWorld is highly compute-intensive and time-consuming to run, we perform a total of 60 training steps for RL training, using $n_\mathcal{T} = 2$ and $n_\pi = 8$. All training are performed on top of Qwen2.5-VL-7B-Instruct and Qwen2.5-VL-32B-Instruct \citep{bai2025qwen25vltechnicalreport} using 16xH100, denoted as ``\Distill{}-7B'' and ``\Distill{}-32B'' in \Cref{tbl:android_rl_result}, respectively.

\subsection{Other Implementation/Evaluation Details}
\label{subsec:Implementation/Evaluation Details}

In this work, we focus on end-to-end training (SFT + RL), and hence selected VLMs capable of directly interacting with android's GUI interface.
This include models such as Qwen2.5-VL \citep{bai2025qwen25vltechnicalreport} and UI-Tars \citep{qin2025uitarspioneeringautomatedgui}.
While these models have undergone specific finetuning on mobile control tasks, at the time of the work we were unable to find evaluation scripts that supports using these models on AndroidWorld.
To our best effort, we utilized the official mobile-use prompts provided by the respective repositories, as well as prompts from recent work such as \citep{gou2025navigatingdigitalworldhumans}. 
However, we were unable to fully reproduce the reported performance, especially for UI-Tars 1.5.
At the time of this work, we find similar concerns has also been raised publicly (e.g., \url{https://github.com/bytedance/UI-TARS/issues/83}, \url{https://github.com/UI-Tars/UI-Tars/issues/155}, \url{https://github.com/UI-Tars/UI-Tars/issues/121}).
To this end, we focus on using Qwen2.5-VL for consistency with other experiments conducted in the rest of the paper.

\begin{table}
\caption{Prompt used by \wmrolloutshort{} to refine the agent's original response given actual next-state information. The next-state information is obtained by 1) extracting the final chosen plan from the agent's response (e.g., left, left, up in Sokoban), and 2) executing the plan in the environment to obtain the actual next states.}
\label{tab:wmrefine_prompt}
\begin{tabular}{p{13.5cm}}
\toprule
Prompt \\
\midrule
\emph{\textcolor{gray}{// ...omitting some text}}
\newline
\# Current observation
\newline
\textcolor{blue}{\{}\textcolor{blue}{current\_observation}\textcolor{blue}{\}}
\newline

\# Example response and feedback
\newline
To help you reason and plan better, we have explored some plans for the current step and obtained the following feedback from the environment:
\newline
\#\# Example response
\newline
\textcolor{blue}{\{}\textcolor{blue}{agent\_original\_response}\textcolor{blue}{\}}
\newline
\#\# Ground truth feedback
\newline
\textcolor{blue}{\{}\textcolor{blue}{actual\_next\_observations\_after\_executing\_agent's\_plan}\textcolor{blue}{\}}
\newline

\# Back to the current step
\newline
Now, the environment has been reset back to the current observation/current step. It's your turn to refine the example response based on the ground truth feedback. You should think about:
\newline
- Correctness: is the example response aligned with the feedback? did the feedback reveal some incorrect/ineffective actions in the example response?
\newline
- Progress: did the the environment feedback show positive progress towards solving the task?
Note: the example response may hallucinate incorrect outcomes different from the ground truth feedback. You should avoid coming up with similar hallucinations in your response.
\newline

If you think the example response is correct and has made progress, no revision is needed and your should **directly output the example response verbatim**.
\newline
Otherwise, you should modify the example response's thinking process/plan/action to be consistent with the environment feedback. Specifically, you should:
\newline
1. **Incorporate all relevant details from the feedback** into the example response and then **improve its accuracy and progress**. Be detailed when adding information from the feedback into the response.
\newline
2. The final refined response should **read as a standalone answer, without saying its a revision or referencing the feedback section**. The output should look like the final authoritative response that faithfully incorporated relevant details from the feedback.
\newline

Now, refine the example response. Keep the same tone and format as the example response. Ensure the final plan/action is logically consistent with your (refined) thinking process. DO NOT generate anything else.
\\
\bottomrule
\end{tabular}
\end{table}
\begin{table}
\caption{Example state in Sokoban and our model's generated action}
\label{tab:sokoban_example}
\begin{tabular}{p{1.5cm} p{11.5cm}}
\toprule
Type & Content \\
\midrule
Input & \emph{\textcolor{gray}{// ...omitting some text about sokoban game rules}}
\newline
\# Current Step
\newline
Your current observation is:
\newline
{    Col 0    Col 1    Col 2    Col 3    Col 4    Col 5}
\newline
{Row 0    \#     \#     \#     \#     \#     \#    }
\newline
{Row 1    \#     \_     \_     \#     \#     \#    }
\newline
{Row 2    \#     \_     P     \#     \#     \#    }
\newline
{Row 3    \#     \_     X     \_     \_     \#    }
\newline
{Row 4    \#     \_     \_     O     \_     \#    }
\newline
{Row 5    \#     \#     \#     \#     \#     \#    }
\newline

Currently, the player is at (2, 2); boxes are at (3, 2); unsolved targets are at (4, 3).
\newline
Next to the player, there is a floor above the player at (1, 2), a box below the player at (3, 2), a floor left of the player at (2, 1), a wall right of the player at (2, 3).
\newline
Your admissible actions are ["up", "down", "left", "right"].
\newline

Now it's your turn to make a move (choose ONE action only for the current step).
\newline
You should first reason step-by-step about the current situation — observe the positions of boxes and targets, plan a path to push a box toward a target, and avoid traps like corners or walls. This reasoning process MUST be enclosed within <think> </think> tags.
\newline
Once you've finished your reasoning, you should choose a plan as well as an admissible action for current step. The plan needs to be presented as a sequence of admissible actions within <plan> </plan> tags (e.g., <plan>up, right, up</plan>). The action needs to be presented within <action> </action> tags (e.g., <action>up</action>).
\\
\cmidrule{2-2}
Output & <think>Okay, let me look at the current situation. The player is at (2, 2), with a box at (3, 2) and an unsolved target at (4, 3). Let's think about a few possible plans.
\newline

Maybe we can try moving "down". This action will lead to the state where the player is at (3, 2), pushing the box to (4, 2). Is this a good next step? Let's think:
\newline
- After "down", maybe do "left". This results in the player moving to (3, 1), with the box remaining at (4, 2) and the target at (4, 3). After that, maybe "down", which leads to the player moving to (4, 1), with the box still at (4, 2) and the target at (4, 3). The player now has a box to the right and a floor above. Discounted success rate if continued further: around 80\%.
\newline
- After "down", maybe do "left". This results in the player moving to (3, 1) with the box still at (4, 2). After that, maybe "down", which leads to the player moving to (4, 1) with the box and target unchanged. Next, try "right", which moves the player to (4, 2) and pushes the box onto the target at (4, 3), solving it. Discounted success rate if continued further: around 100\%.
\newline

Based on these simulations, "down" is the best action for the current step. This is because the second plan (down -> left -> down -> right) achieves a discounted success rate of 100\% in 4 steps.
\newline
</think>
\newline

<plan>down, left, down, right</plan>
\newline
<action>down</action>
\\
\bottomrule
\end{tabular}
\end{table}
\begin{table}
\caption{Example state in ALFWorld and our model's generated action}
\label{tab:alfworld_example}
\begin{tabular}{p{1.5cm} p{11.5cm}}
\toprule
Type & Content \\
\midrule
Input & \emph{\textcolor{gray}{// ...omitting some text about alfworld rules}}
\newline
Your task is to: put some book on sidetable.
\newline
Your admissible actions of the current situation are: ['go to bed 1'
 'go to diningtable 1'
 'go to drawer 1'
 'go to drawer 2'
 'go to drawer 3'
 'go to drawer 4'
 'go to garbagecan 1'
 'go to sidetable 1'
 'go to sidetable 2'
 'inventory'
 'look'].
\newline

Now it's your turn to take an action.
\newline
You should first reason step-by-step about the current situation, and think about plans to explore/solve the task. To plan well, you should think about the plan's future outcomes and whether it is making progress or not. This reasoning process MUST be enclosed within <think> </think> tags.
\newline
Once you've finished your reasoning, you should choose a plan as well as an admissible action for current step. The plan needs to be presented as a sequence of admissible actions within <plan> </plan> tags (e.g., <plan>go to desk 1, take mug 1 from desk 1</plan>). The action needs to be presented within <action> </action> tags (e.g., <action>go to desk 1</action>).
\\
\cmidrule{2-2}
Output & <think>Okay, let me look at the current situation. Currently, I am in the middle of a room with several objects around me, including two sidetables (sidetable 1 and sidetable 2), where my task is to put a book on one of them. Let's think about a few possible plans.
\newline

Maybe we can try Go to diningtable 1. This action may lead to arriving at diningtable 1, where I see a creditcard 1, a book 3, a cdplayer 2, a laptop, a pencil, three pens, a nutshell, and a vase. Is this a good next step? Let's think:
\newline
- After Go to diningtable 1, maybe I can take book 3. This results in picking up the book 3 from the diningtable. Discounted success rate if continued further: around 80\%.
\newline
- After Go to diningtable 1, maybe I can inventory myself. This may reveal that nothing is being carried, so taking anything is impossible. Discounted success rate if continued further: around 10\%.
\newline

Alternatively, maybe we can try Open drawer 1. This action may lead to opening drawer 1 and seeing nothing inside. Is this a good next step? Let's think:
\newline
- After Open drawer 1, maybe "examine drawer 2" reveals a combination lock and a keychain inside. Discounted success rate if continued further: around 40\%.
\newline

Based on these simulations, going to the diningtable has the highest discounted success rate (80\%) as book 3 may be found. This achieves the highest success if followed by "take book 3", achieving a 80\% success rate.
\newline
</think>
\newline

<plan>go to diningtable 1, take book 3</plan>
\newline
<action>go to diningtable 1</action>
\\
\bottomrule
\end{tabular}
\end{table}
\begin{table}
\caption{Sokoban prompt to extract plan and imagined observation from \textcolor{blue}{an agent's response}}
\label{tab:sokoban_extract_prompt}
\begin{tabular}{p{13.5cm}}
\toprule
Prompt \\
\midrule
\emph{\textcolor{gray}{// ...omitting some text about sokoban game rules}}
\newline
\# Extraction/parsing rules
\newline
Your task is to parse the response and extract the following information, IF present.
\newline
1) simulation branches
\newline
  - definition: one sequence of actions the agent planned to solve the puzzle
\newline
  - example: \emph{\textcolor{gray}{// ...omitting some text }}
2) discounted success rates
\newline
  - definition: the (discounted) success rate of the simulation branch, or some numeric evaluation of how much progress that branch makes towards the goal.
\newline
  - example: \emph{\textcolor{gray}{// ...omitting some text }}
\newline
3) final chosen branch
\newline
  - definition: the simulation branch/plan that caused the agent's final decision for the current step.
\newline
  - example: Based on these simulations, "up" is the best action for the current step. This is because after "up", the player can proceed with "left" and "up" again, which achieves a discounted success rate of around 90\% in 3 steps.
\newline
  - example output: ["up", "left", "up"]
\newline
  - note: The agent chose "up" as the next action. However, we need to find the ENTIRE branch/plan that caused the agent's current decision, which is ["up", "right", "down"] in this case.
\newline
  - note: if the agent did not explicitly mention which branch is chosen, you should choose the branch in the response with the highest discounted success rate.
\newline
4) final imagined observation
\newline
  - definition: the imagined observation after executing the final chosen branch.
\newline
  - example: After "up", "left", "up", the player pushed the box to (4,4). Now, the player is at (4, 3), with the box on target below at (4, 4). The player has a floor above at (2, 4)... The target is ... This is the best branch according to the discounted success rate. So the next action should be "up".
\newline
  - example output: The player pushed the box to (4,4). Now, the player is at (4, 3), with the box on target below at (4, 4). The player has a floor above at (2, 4)... The target is ...
\newline
  - note: DO NOT include the action sequence in this field. Only keep the description of the player/boxes/targets/walls position AFTER the last action in the final chosen branch.
\newline
  - note: \emph{\textcolor{gray}{// ...omitting some text }}
\newline

\# Your task
\newline
Your task is to output a JSON object in the following format:
\newline
<json>
\newline
\{
\newline
    "extracted\_branches": [
        ...\emph{\textcolor{gray}{// ...omitting some text }}
    ],
\newline
    "extracted\_final\_chosen\_branch": \{
\newline
        "actions": ["action 1", "action 2", ..., "action n"], \# the ENTIRE branch/plan that caused the agent's current decision
\newline
        "last\_observation": "detailed, comprehensive description of the imagined observation AFTER executing the entire action sequence above.",
\newline
        "discounted\_success\_rate": ...(a number between 0 to 100. -1 if the agent did not mention the discounted success rate)
\newline
    \}
\newline
\}
\newline
</json>
\newline

\# Input response
\newline
\textcolor{blue}{\{}\textcolor{blue}{input\_agent\_response}\textcolor{blue}{\}}
\newline

\# Your task
\newline
Now, parse the response and output the JSON object enclosed by <json> and </json> tags. DO NOT generate anything else.
\\
\bottomrule
\end{tabular}
\end{table}

\begin{table}
\caption{Sokoban prompt to evaluate the quality of the next-states imagined by an agent in its reasoning process, using the actual next-states as references.}
\label{tab:sokoban_judge_prompt}
\begin{tabular}{p{13.5cm}}
\toprule
Prompt \\
\midrule
\emph{\textcolor{gray}{// ...omitting some text about sokoban game rules}}
\newline
\# Evaluation rules
\newline
Provide an overall score between 0.0 and 1.0 based on the following two dimensions.
Start with a score of 0.0, and add points to the score if the criteria are satisfied. Add 0.0 if a criteria is not satified. DO NOT deduct points if a criteria is not satified.
\newline
1) correctness (max 0.3 points. if exceeds 0.3, cap it at 0.3)
\newline
  - in the imagination description, the coordinates of the player are correct; add 0.1 point
  \newline
  - in the imagination description, some of the mentioned boxes and targets have correct coordinates; add 0.05 point
  \newline
  - in the imagination description, all mentioned boxes and targets have correct coordinates; add 0.1 point
  \newline
  - in the imagination description, all mentioned walls and empty spaces have correct coordinates; add 0.05 point
  \newline
2) progress (max 0.7 points. if exceeds 0.7, cap it at 0.7)
  \newline
  - in the reference observation, if the task is completely solved (all boxes are on targets); add 0.7 point
  \newline
  - relative to the current observation, if the reference observation shows major progress (unsolved boxes are moved much closer to targets, task close to be solved); add 0.5 point
  \newline
  - relative to the current observation, if the reference observation shows minor progress (unsolved boxes are moved a bit closer to targets); add 0.1-0.3 point, depending on how much progress is shown
  \newline
  - relative to the current observation, if the reference observation shows no meaningful progress; assign 0.0 point for this dimension
  \newline
  - in the reference observation, if the task is no longer solvable (e.g., one of the boxes is pushed into a corner and cannot be moved anymore); assign 0.0 point for this dimension
  \newline
\emph{\textcolor{gray}{// ...omitting some text }}
\newline

\# Your output format
\newline
Your task is to output a JSON object in the following format:
\newline
<json>
\newline
\{
\newline
    "correctness analysis": "which correctness criteria in the evaluation rules are satisfied, and which are not.", \# no more than 50 words
\newline
    "correctness score": 0.0-0.3, \# score for the correctness dimension
\newline
    "progress analysis": "which progress criteria in the evaluation rules are satisfied, and which are not.", \# no more than 50 words
\newline
    "progress score": 0.0-0.7, \# score for the progress dimension
\newline
    "score": 0.0-1.0 \# total score; add the correctness score and progress score
\newline
\}
\newline
</json>
\newline

\# Current observation
\newline
\textcolor{blue}{\{}\textcolor{blue}{current\_obs}\textcolor{blue}{\}}

\# Agent imagined observation after some actions
\newline
\textcolor{blue}{\{}\textcolor{blue}{agent\_imagined\_next\_actions\_and\_obs}\textcolor{blue}{\}}

\# Reference observation after some actions
\newline
\textcolor{blue}{\{}\textcolor{blue}{actual\_next\_obs}\textcolor{blue}{\}}
\newline

\# Your task
\newline
Now, provide an evaluation analysis and score according to the evaluation rules above. Output the JSON object enclosed by <json> and </json> tags. DO NOT generate anything else.
\\
\bottomrule
\end{tabular}
\end{table}

\begin{table}
\caption{ALFWorld prompt to extract plan and imagined observation from \textcolor{blue}{an agent's response}}
\label{tab:alfworld_extract_prompt}
\begin{tabular}{p{13.5cm}}
\toprule
Prompt \\
\midrule
\emph{\textcolor{gray}{// ...omitting some text about sokoban game rules}}
\newline
\# Extraction/parsing rules
\newline
Your task is to parse the response and extract the following information, IF present.
\emph{\textcolor{gray}{// ...omitting some text }}
\newline
3) final chosen branch
  \newline
  - definition: the simulation branch/plan that caused the agent's final decision for the current step.
  \newline
  - example: Based on these simulations, "go to countertop 1" is the best action for the current step. This is because this followed by "go to countertop 2" leads to a high chance of finding a mug. Therefore, the next action for the current step should be "go to countertop 1".
  \newline
  - example output: ["go to countertop 1", "go to countertop 2"]
  \newline
  - note: The agent chose "go to countertop 1" as the next action. However, we need to find the ENTIRE branch/plan that caused the agent's current decision, which is ["go to countertop 1", "go to countertop 2"] in this case.
  \newline
  - note: if the agent did not explicitly mention which branch is chosen, you should choose the branch in the response with the highest discounted success rate.
  \newline
4) final imagined observation
  \newline
  - definition: the imagined observation after executing the final chosen branch.
  \newline
  - example: After "go to shelf 1", "take pencil 2 from shelf 1" results in successfully picking up a pencil. This is the best branch according to the discounted success rate. So the next action should be "go to shelf 1".
  \newline
  - example output: The agent successfully picks up a pencil.
  \newline
  - note: DO NOT include the action sequence in this field. Only keep the description of the imagined observation AFTER the last action in the final chosen branch.
  \newline
  - note: In general, you should gather the most comprehensive and detailed description found in the response (i.e., especially try to include any mention of what objects is present). If this description is scattered across multiple places in the response, MERGE them into a single, continuous description.
  \newline

\# Your task
\newline
Your task is to output a JSON object in the following format:
<json>
\newline
\{
\newline
    "extracted\_branches": [
        ...\emph{\textcolor{gray}{// ...omitting some text }}
    ],
    \newline
    "extracted\_final\_chosen\_branch": \{
        \newline
        "actions": ["action 1", "action 2", ..., "action n"], \# the ENTIRE branch/plan that caused the agent's current decision
        \newline
        "last\_observation": "detailed, comprehensive description of the imagined observation AFTER executing the entire action sequence above.",
        \newline
        "discounted\_success\_rate": ...(a number between 0 to 100. -1 if the agent did not mention the discounted success rate)
        \newline
    \}
\}
\newline
</json>
\newline

\# Input response
\newline
\textcolor{blue}{\{}\textcolor{blue}{input\_agent\_response}\textcolor{blue}{\}}
\newline

\# Your task
\newline
Now, parse the response and output the JSON object enclosed by <json> and </json> tags. DO NOT generate anything else.
\\
\bottomrule
\end{tabular}
\end{table}

\begin{table}
\caption{ALFWorld prompt to evaluate the quality of the next-states imagined by an agent in its reasoning process, using the actual next-states as references.}
\label{tab:alfworld_judge_prompt}
\begin{tabular}{p{13.5cm}}
\toprule
Prompt \\
\midrule
\emph{\textcolor{gray}{// ...omitting some text about sokoban game rules}}
\newline
\# Evaluation rules
\newline
Provide an overall score between 0.0 and 1.0 based on the following two dimensions.
\newline
1) correctness (max 0.3 points. if exceeds 0.3, cap it at 0.3)
  \newline
  - in the imagined observation, it is near identical to the reference observation; add 0.3 point
  \newline
  - in the imagined observation, key object(s) required by the goal are found, and they are also present in the reference observation; add 0.2 point
  \newline
  - in the imagined observation, relevant location(s) required by the goal are visited, and the description is somewhat aligned with the reference observation; add 0.1-0.2 point, depending on how much the description is aligned with the reference observation.
  \newline
  - in the imagined observation, key object(s) required by the goal are found, but these key object(s) are *NOT* present in the reference observation; assign 0.0 point
  \newline
  - in the reference observation, it shows nothing happened; directly assign 0.0 point for this dimension
  \newline
2) progress (max 0.7 points. if exceeds 0.7, cap it at 0.7)
  \newline
  - in the reference observation, if the goal is completely solved (all required items are found/moved/heated/etc to or at the correct location, goal is achieved); add 0.7 point
  \newline
  - relative to the current observation and action history, if the reference observation shows major progress (i.e., objects required by the goal are found); add 0.5 point
  \newline
  - relative to the current observation and action history, if the reference observation shows minor progress (i.e., objects related to the goal are found, or locations relevant to the goal are visited); add 0.1-0.3 point, depending on *how useful this information is, beyond what was already known in the current state and action history*.
  \newline
  - relative to the current observation and action history, if the reference observation shows no meaningful progress (nothing happened); assign 0.0 point for this dimension
  \newline
\emph{\textcolor{gray}{// ...omitting some text }}
\newline

\# Your output format
\newline
Your task is to output a JSON object in the following format:
<json>
\newline
\{
    \newline
    "correctness analysis": "...", \# no more than 50 words
    \newline
    "correctness score": 0.0-0.3, \# score for the correctness dimension
    \newline
    "progress analysis": "...", \# no more than 50 words
    \newline
    "progress score": 0.0-0.7, \# score for the progress dimension
    \newline
    "score": 0.0-1.0 \# total score; add the correctness score and progress score
    \newline
\}
\newline
</json>
\newline

\# Action history
\newline
The current goal is to: \textcolor{blue}{\{}\textcolor{blue}{task\_description}\textcolor{blue}{\}}
\newline
\textcolor{blue}{\{}\textcolor{blue}{action\_history}\textcolor{blue}{\}}

\# Current observation
\newline
\textcolor{blue}{\{}\textcolor{blue}{current\_obs}\textcolor{blue}{\}}

\# Agent imagined observation after some actions
\newline
\textcolor{blue}{\{}\textcolor{blue}{agent\_imagined\_next\_actions\_and\_obs}\textcolor{blue}{\}}

\# Reference observation after some actions
\newline
\textcolor{blue}{\{}\textcolor{blue}{actual\_next\_obs}\textcolor{blue}{\}}
\newline

\# Your task
\newline
Now, provide an evaluation analysis and score according to the evaluation rules above. Output the JSON object enclosed by <json> and </json> tags. DO NOT generate anything else.
\\
\bottomrule
\end{tabular}
\end{table}

\end{document}